\PassOptionsToPackage{table}{xcolor}
\documentclass{article}
\usepackage{array} 
\usepackage{multirow}
\usepackage{xcolor}

\usepackage{microtype}
\usepackage{graphicx}
\usepackage{subcaption}
\usepackage{booktabs} 

\usepackage{hyperref}

\usepackage[accepted]{icml2026}

\usepackage{amsmath}
\usepackage{amssymb}
\usepackage{mathtools}
\usepackage{amsthm}

\usepackage[capitalize,noabbrev]{cleveref}

\theoremstyle{plain}

\theoremstyle{definition}

\theoremstyle{remark}

\usepackage{enumitem} 
\usepackage[table]{xcolor}
\usepackage{algorithm}
\usepackage{algorithmic}
\usepackage{thmtools, thm-restate}
\usepackage{bbm} 

\definecolor{marineblue}{RGB}{0,114,189}
\newcommand{\dearcode}{\href{https://github.com/dahyedahye/dear/}{\textcolor{marineblue}{\texttt{github.com/dahyedahye/dear}}}}
\icmltitlerunning{Dissect and Prune: Enhancing Robustness in AI-Generated Image Detection}

\begin{document}

\twocolumn[
  \icmltitle{Dissect and Prune: Enhancing Robustness in AI-Generated Image Detection}



  \icmlsetsymbol{equal}{*}

  \begin{icmlauthorlist}
    \icmlauthor{Dahye\hspace{0.23em}Kim}{aisi}
    \icmlauthor{Jaehyun\hspace{0.23em}Choi}{aisi}
    \icmlauthor{Hyun\hspace{0.23em}Seok\hspace{0.23em}Seong}{aisi,skku}
    \icmlauthor{Seongho\hspace{0.23em}Kim}{aisi}
    \icmlauthor{Donghun\hspace{0.23em}Lee}{aisi}
    \icmlauthor{Sungwon\hspace{0.23em}Yi}{aisi}
    \icmlauthor{Jang-Ho\hspace{0.23em}Choi}{aisi}
  \end{icmlauthorlist}

  \icmlaffiliation{aisi}{Korea AI Safety Institute (AISI), ETRI, Seongnam, South Korea}
  \icmlaffiliation{skku}{Department of Artificial Intelligence, Sungkyunkwan University, Suwon, South Korea}

  \icmlcorrespondingauthor{Jang-Ho Choi}{janghochoi@etri.re.kr}

  \icmlkeywords{Machine Learning, ICML}

  \vskip 0.3in
]



\printAffiliationsAndNotice{}  

\begin{abstract}
    While existing AI-generated image detectors report high performance, we identify that this is largely driven by a critical \emph{prediction asymmetry}: a bias toward the real class that severely limits sensitivity to generated content, especially under standard post-processing operations such as compression and resizing. We hypothesize that this stems from the model's reliance on spurious features, distracting signals that obscure true generative artifacts. To address this, we propose DEAR (Dissect and Prune), which leverages inpainted images to identify and prune these interfering components. Specifically, we find that features strongly aligned to either inpainted or non-inpainted regions are less robust to post-processing. By measuring the alignment between channel activations and inpaint masks, DEAR removes features at both extremes, retaining only those that capture genuine generative artifacts. Experimental results demonstrate that our approach significantly enhances robustness against unseen generators and post-processing, effectively mitigating the prediction asymmetry. Our code is available at \dearcode.
\end{abstract}

\section{Introduction}

\begin{figure}[t]
    \centering
    \includegraphics[width=0.99\linewidth]{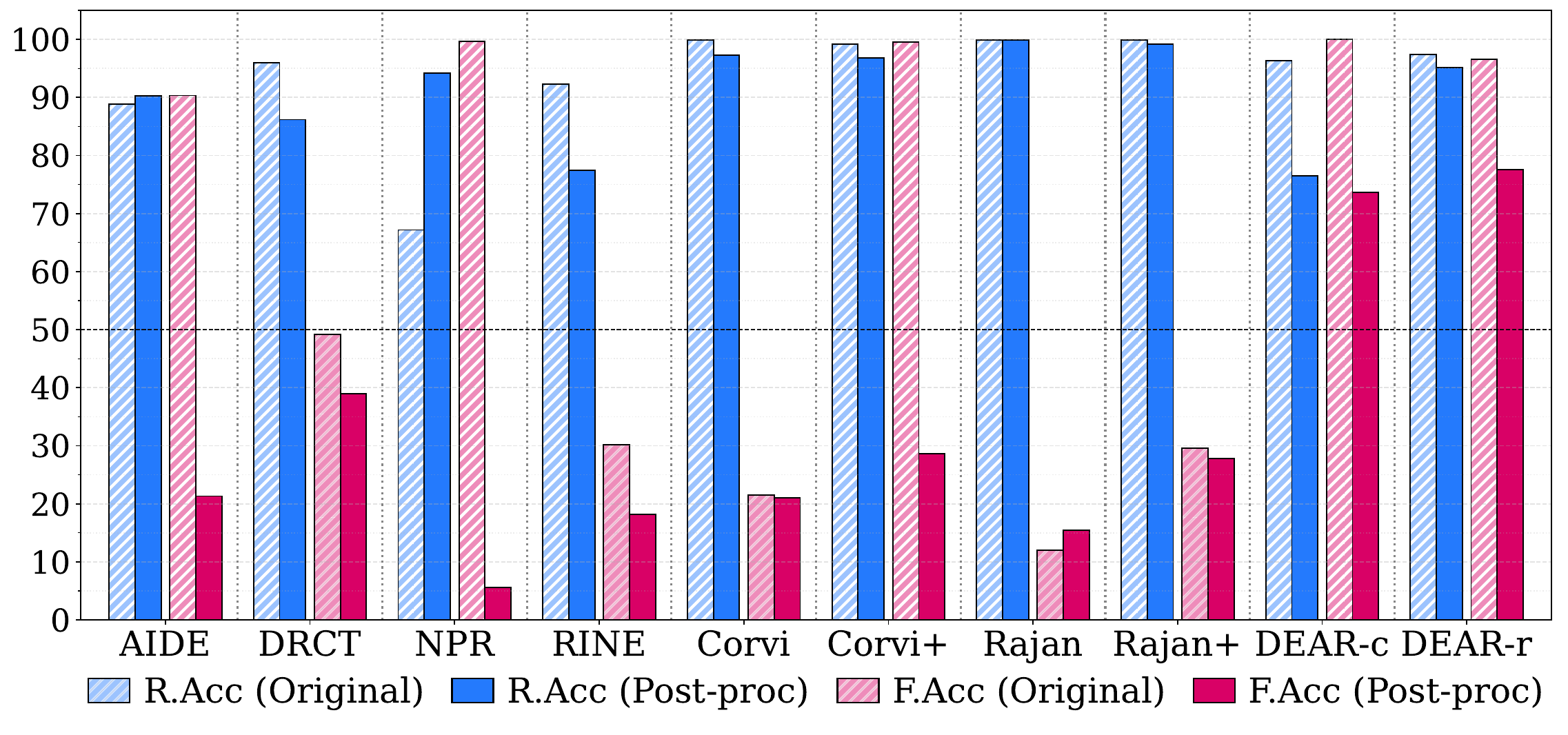}
    \caption{\textbf{Prediction Asymmetry in AIGI Detection.} Comparison of real accuracy (R.Acc) and fake accuracy (F.Acc) before and after post-processing on FLUX~\cite{flux} generated images. Existing detectors maintain high R.Acc regardless of post-processing, but F.Acc drops dramatically after post-processing is applied. This asymmetric degradation reveals that detectors rely on fragile spurious features for fake detection. DEAR mitigates this asymmetry by pruning such features and retaining only robust forensic signals.}
    \label{fig:asymmetric_robustness}
    \vspace{-7px}
\end{figure}

Recent advances in generative models, particularly diffusion~\cite{ho2020denoising, rombach2022high} and flow-based models~\cite{lipman2022flow}, have enabled the synthesis of near-photorealistic images that are increasingly difficult to distinguish from real photographs. While these technologies offer remarkable generative capabilities, they also raise significant societal concerns, including the proliferation of misinformation through convincing fake imagery, the creation of non-consensual deepfakes, and potential copyright infringement. As AI-generated images (AIGI) become more prevalent and sophisticated, the development of reliable detection methods has become an urgent and critical challenge.

To address this challenge, numerous AIGI detectors have been proposed, spanning CNN-based approaches~\cite{wang2020cnn,corvi2023detection,tan2024rethinking,rajan2025aligned,rajan2025stay,chen2024drct}, CLIP-based methods~\cite{ojha2023towards,cozzolino2024raising,yan2025effort,tan2025c2p}, Vision Transformer-based methods~\cite{guillaro2025bias,chen2025dual}, and frequency-domain analysis techniques~\cite{tan2024frequency}. These methods report impressive performance on benchmark datasets, often achieving high accuracy or average precision. Such results have suggested that AIGI detection is approaching a solved problem, with detectors capable of reliably distinguishing AI-generated from real images.

However, upon closer examination of detector performance, we observe a critical phenomenon that we term \textit{prediction asymmetry}: a systematic bias where high overall accuracy is sustained primarily by the near-perfect recognition of real images, while the sensitivity to AI-generated content remains disproportionately low. This imbalance obscures the detector's actual reliability, as high overall metrics hide a failure to identify the very target class they are designed to detect. This behavior becomes markedly more pronounced under common real-world post-processing such as JPEG compression and resizing~\cite{li2025artificial}. Under such conditions, the accuracy on fake images decreases significantly as the detector gravitates toward the real class, becoming a trivial classifier that defaults to predicting real images.


We hypothesize that this prediction asymmetry stems from the detector's reliance on spurious correlations rather than robust forensic traces~\cite{rajan2025stay}. Existing detectors often exploit dataset-specific biases, such as compression artifacts, as indicators of realness, while simultaneously overfitting to fragile shortcuts specific to generated content~\cite{rajan2025stay,grommelt2024fake,yan2025effort,kashiani2025freqdebias,ma2025specificity}. This reliance leads to failure even in the original setting. When encountering unseen generators that lack the specific fingerprints seen during training, the detector fails to recognize the generated content~\cite{li2025artificial}. This vulnerability is further exacerbated by post-processing. Since these non-intrinsic signals are sensitive to perturbations, their degradation diminishes the discriminative cues available to the detector. Consequently, the model exhibits a systematic shift toward predicting the real class.

To address the prediction asymmetry driven by spurious correlations, we aim to identify and mitigate the influence of specific detector features responsible for these non-robust dependencies. To this end, we draw upon the principles of Network Dissection~\cite{bau2017network, bau2018gan,tousi2021automatic}, a framework designed to interpret internal representations. Successfully dissecting a forensic detector, however, requires a precise ground truth to distinguish between features activating on generated artifacts and those responding to authentic signals. We identify \textit{inpainted images} as the optimal probe for this purpose. Since these images spatially isolate AI-generated content within inpainted regions from the authentic context in non-inpainted backgrounds while offering exact ground truth masks, they provide the necessary reference to rigorously correlate feature activations with pixel provenance.

Building on this insight, we propose DEAR (\textbf{D}iss\textbf{E}ct \textbf{A}nd P\textbf{R}une) for robust AIGI detection. Our approach adopts a selective pruning strategy. First, we employ inpainted images as a diagnostic tool, creating a controlled environment where generated pixels coexist with real image contexts. Second, we perform detector dissection by quantifying the divergence in feature activations between the generated (inpainted) regions and the real (non-inpainted) backgrounds. This metric allows us to determine whether a feature is primarily driven by generative artifacts or authentic signals. Third, we refine the classifier by pruning features at both extremes of this spectrum, targeting channels strongly aligned with generated content and those biased toward real backgrounds. We find that features at these extremes are susceptible to degradation. Consequently, removing these components forces the detector to rely on robust forensic signals, effectively eliminating the spurious representations that degrade performance under perturbations.

Our main contributions are as follows:
\textbf{(1)} We identify prediction asymmetry as a fundamental limitation of current AIGI detectors, where high overall accuracy conceals a systematic bias toward the real class that severely degrades fake detection under perturbations.
\textbf{(2)} We propose DEAR, a dissection guided feature pruning mechanism that leverages inpainted images to measure feature alignment and selectively removes spurious features at both extremes, retaining only those that capture genuine generative artifacts.
\textbf{(3)} We demonstrate that DEAR significantly enhances robustness against unseen generators and post-processing operations through extensive experiments, effectively mitigating prediction asymmetry.

\section{Preliminaries}
\begin{figure*}[t]
  \centering
  \includegraphics[width=\textwidth]{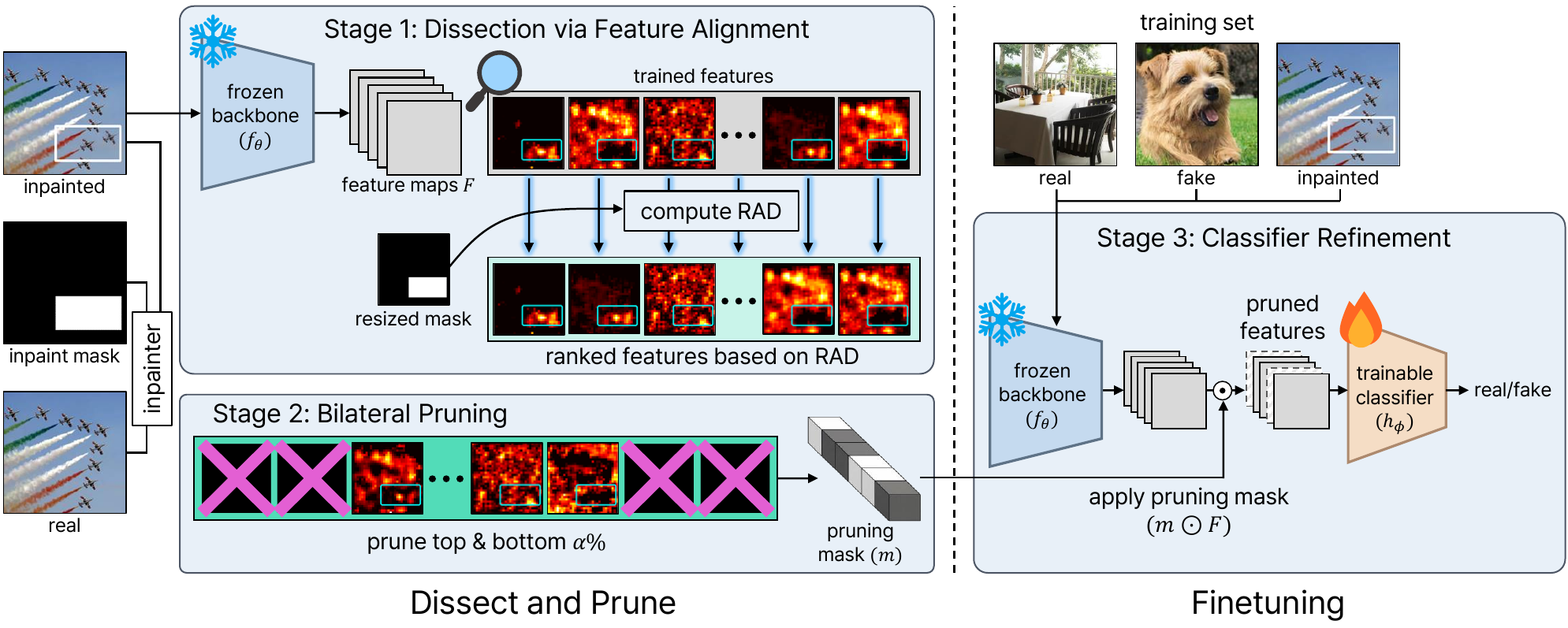}
    \caption{\textbf{Overview of Dissect and Prune (DEAR).} Our method operates in three stages: (1) \textbf{Dissection via Feature Alignment}, where we use inpainted images and masks as a diagnostic tool to measure how strongly each feature channel aligns with generated or real regions through Regional Activation Discrepancy (RAD); (2) \textbf{Bilateral Pruning}, where we identify and remove channels at both extremes of the RAD distribution, as these represent non-intrinsic components relying on fragile shortcuts or spurious realness indicators; and (3) \textbf{Classifier Refinement}, where we fine-tune the final linear classifier on the pruned feature space obtained by Dissect and Prune (DEAR) while the backbone remains frozen. (Section~\ref{sec:method}).}
    
  \label{fig:main_fig}
\end{figure*}

To systematically evaluate the reliability of detectors, we analyze their behavior across diverse generators and perturbations, revealing a critical failure mode which we term \emph{Prediction Asymmetry}: a systematic bias where detectors maintain high accuracy on real images while failing to identify AI-generated images. This phenomenon aligns with findings in recent large-scale benchmarks like AIGIBench~\cite{li2025artificial}, which we analyze along two dimensions: generalization to unseen generators and robustness under post-processing.

\subsection{Motivation: Prediction Asymmetry Problem}
\label{sec:asymmetric}

First, we observe a pronounced generalization gap under unseen-generator shift. As illustrated in Figure~\ref{fig:asymmetric_robustness}, Corvi achieves near-perfect R.Acc (99.9\%) on original FLUX~\cite{flux} images but a markedly lower F.Acc of 21.5\%, indicating that detectors default to the real class when specific training fingerprints are absent. Second, this bias amplifies into a robustness collapse under post-processing. For instance, NPR's average F.Acc plummets from 95.9\% to 12.2\% while its R.Acc rises from 67.2\% to 94.2\%, reducing it to a trivial classifier that labels all inputs as real once fragile cues are disrupted. This susceptibility exposes a critical practical limitation where aggregate metrics like AUC mask a structural bias toward the real class, inflating reliability assessments in real-world settings.

We attribute this asymmetry to the detector's reliance on \emph{spurious correlations}, which fall into two failure modes. First, detectors frequently exploit dataset-specific biases, such as JPEG compression artifacts or high-frequency details, as spurious indicators of realness~\cite{grommelt2024fake,rajan2025stay}. Consequently, when perturbations introduce similar artifacts to fake images, detectors misclassify them as real. Second, regarding spurious features specific to generated content, evidence suggests that detectors overfit to monotonous and non-robust artifacts, as exemplified by low-rank traces~\cite{yan2025effort}, spectral biases~\cite{kashiani2025freqdebias}, and generator-specific cues~\cite{ma2025specificity}. These artifacts serve as fragile shortcuts~\cite{geirhos2020shortcut} that are obliterated by post-processing, leaving the detector to default to the real class.

The confluence of these factors explains the observed prediction asymmetry and mirrors the concept of a \emph{modality gap}~\cite{li2025artificial}, where real images possess concentrated and consistent statistics while AI-generated images rely on artifacts characterized by high modality variance. Motivated by these insights into spurious correlations linked to both real and fake classes, we propose DEAR to explicitly dissect and prune features driven by these spurious dependencies, thereby isolating robust forensic features. Empirical evidence for both failure modes is provided in Appendix~\ref{app:spurious_evidence}.

\subsection{Network Dissection for Feature Analysis}
\label{sec:prelim_dissect}
To address this asymmetry, it is critical to identify which components of the feature backbone are responsible for encoding spurious correlations. Since deep neural networks operate as black boxes, we require a quantitative framework to interpret their internal representations. To this end, we adopt \textbf{Network Dissection}~\cite{bau2017network,bau2018gan}.

The central premise of this framework is that individual channels within a feature map often emerge as concept-specific units, activating in response to distinct visual attributes. Formally, for a given unit $u$, its activation map is upsampled to match the image resolution and binarized into a segmentation proposal $\hat{M}_u(\mathbf{x})$ using a statistical threshold. The interpretability of the unit for a concept $c$ is then computed via the Intersection over Union (IoU) score:
\begin{equation}
    \text{IoU}_{u,c} = \frac{\sum_{\mathbf{x} \in \mathcal{D}} |\hat{M}_u(\mathbf{x}) \cap M_c(\mathbf{x})|}{\sum_{\mathbf{x} \in \mathcal{D}} |\hat{M}_u(\mathbf{x}) \cup M_c(\mathbf{x})|},
\end{equation}
where $M_c(\mathbf{x})$ denotes the ground truth mask for concept $c$.

While the original framework measures alignment with high-level semantic concepts (e.g., objects, textures), we extend this approach to the problem of AI-generated image detection. In Section~\ref{sec:method}, we will adapt this formulation to measure feature alignment with both \textit{generative artifacts} (inpainted regions) and \textit{authentic signals} (backgrounds). This approach enables us to profile feature behaviors based on their sensitivity to generated versus authentic content, allowing us to categorize features that strictly activate on fake regions, those tracking real backgrounds, or those exhibiting no distinct regional preference.
\section{Dissect and Prune (DEAR)}
\label{sec:method}
We present \emph{Dissect and Prune} (DEAR), a feature selection framework designed to address the prediction asymmetry problem in AIGI detection. Figure~\ref{fig:main_fig} illustrates an overview of our proposed method. Our approach comprises three stages. We first construct diagnostic inpainted images that contain both real and AI-generated regions. We then dissect detector representations by measuring channel alignment with the inpainted regions. Finally, we prune channels with extreme alignment and refine the linear classifier on the resulting pruned feature representation.



\subsection{Inpainting-Based Diagnostic Data Generation}
\label{sec:inpaint_gen}


To dissect detector representations and isolate features responsible for spurious correlations, we require a diagnostic setting that spatially separates generative artifacts from authentic signals. Inpainted images provide precisely this controlled environment. By replacing a masked region of a real image with content synthesized by a generative model, we create samples where pixel provenance varies across defined spatial boundaries. Crucially, this setup allows us to project features onto a quantitative alignment spectrum. By analyzing this spectrum, we can diagnose which subsets of features are the primary drivers of non-robust behaviors, including those strictly adhering to generated regions, those biased toward authentic backgrounds, or those showing no clear regional preference.

We generate our diagnostic dataset using the inpainting variant of Stable Diffusion 1.5~\cite{rombach2022high}. For each real image $\mathbf{x}_{\text{real}}$, we sample a randomly positioned rectangular binary mask $\mathbf{M} \in \{0, 1\}^{H \times W}$ to define the region for manipulation. The inpainting model then synthesizes content $\mathbf{x}_{\text{gen}}$ for this masked area while conditioning on the surrounding authentic context. To prevent the detector from exploiting trivial edge discontinuities at the mask boundaries, we apply Gaussian blur to the mask edges before compositing. The final inpainted image is constructed as:
\begin{equation}
    \mathbf{x}_{\text{inpaint}} = \mathbf{M} \odot \mathbf{x}_{\text{gen}} + (1 - \mathbf{M}) \odot \mathbf{x}_{\text{real}},
    \label{eq:composite}
\end{equation}
where $\mathbf{x}_{\text{gen}}$ denotes the output from the inpainting model and $\odot$ represents elementwise multiplication. Implementation details are given in Appendix~\ref{app:impl_detail}. Representative examples are provided in Figure~\ref{fig:inpaint_samples} in the Appendix~\ref{app:add_results}. To verify stability across alternative inpainters, we provide an additional ablation in Appendix~\ref{app:rad_robustness}.

A key advantage of this construction is that the ground truth mask $\mathbf{M}$ is known, which naturally provides a precise spatial reference for internal analysis. This naturally facilitates the application of Network Dissection~\cite{bau2017network,bau2018gan}, enabling us to quantitatively measure the alignment between internal feature maps and the generated regions. 

\subsection{Detector Dissection via Feature Alignment}
\label{sec:dissection}

Given inpainted images where the ground truth mask clearly separates generated and real regions, we now propose a method to quantify how individual feature channels align with these regions. Our approach draws inspiration from Network Dissection~\cite{bau2017network}, which identifies interpretable units by measuring overlap between activation maps and semantic concepts. Instead of semantic categories, we measure channel alignment with inpainted regions to expose the reliance on spurious realness indicators and fragile generator shortcuts. By analyzing these alignment characteristics, we identify non-intrinsic components driven by confounding correlations, thereby uncovering the underlying drivers of asymmetric detection.


\paragraph{Regional Activation Discrepancy.}
To quantify the alignment between a feature channel and the inpainted mask, we adopt a density-based metric inspired by the Chan-Vese segmentation model~\cite{chan2001active}, which assumes piecewise constant image regions and measures how well a given contour separates distinct intensity levels.

Let $\mathbf{F}_k \in \mathbb{R}^{h \times w}$ denote the activation map of the $k$th channel from the final convolutional layer, and let $\mathbf{M} \in \{0, 1\}^{h \times w}$ be the downsampled inpaint mask. We define the \emph{Regional Activation Discrepancy} (RAD) as:
\begin{equation}
    S_k = \mu_{\text{in}}^{(k)} - \mu_{\text{bg}}^{(k)},
    \label{eq:rad}
\end{equation}
where
\begin{equation}
    \mu_{\text{in}}^{(k)} = \frac{\sum_{x \in \Omega_{\text{in}}} \mathbf{F}_k(x)}{|\Omega_{\text{in}}|}, \quad
    \mu_{\text{bg}}^{(k)} = \frac{\sum_{x \in \Omega_{\text{bg}}} \mathbf{F}_k(x)}{|\Omega_{\text{bg}}|},
\end{equation}
and $\Omega_{\text{in}} = \{x : \mathbf{M}(x) = 1\}$ denotes the inpainted region and $\Omega_{\text{bg}} = \{x : \mathbf{M}(x) = 0\}$ denotes the background. By normalizing by regional area, RAD remains robust regardless of whether the inpainted region covers a small or large fraction of the image.

Intuitively, a channel with high positive RAD activates strongly within the generated region and weakly in the real background, suggesting sensitivity to generative artifacts. Conversely, a channel with large negative RAD activates preferentially on pixels within real regions.

\begin{figure}[t]
    \centering
    \includegraphics[width=1.0\linewidth]{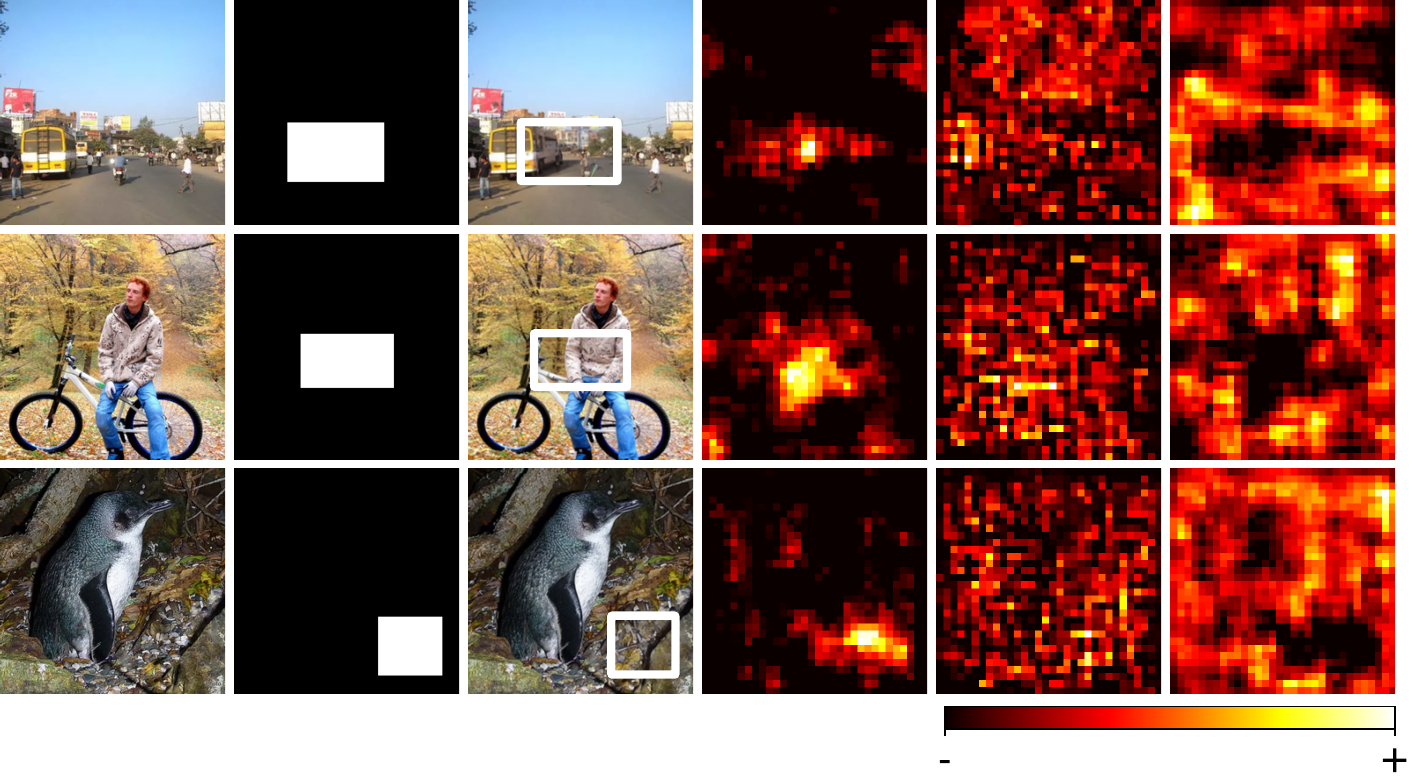}
    \caption{\textbf{Feature Alignment Visualization.} From left to right: original real image, inpainting mask, inpainted image, and activation maps from three representative channels. The high RAD channel (fourth column) activates strongly within the inpainted region, the low RAD channel (sixth column) activates predominantly on the background, and the middle RAD channel (fifth column) shows no clear regional preference.}
    \label{fig:feature_align}
    \vspace{-7px}
\end{figure}

\paragraph{Analysis.}

We apply this dissection procedure to two representative detectors: Corvi~\cite{corvi2023detection} and Rajan~\cite{rajan2025aligned}. Both employ a ResNet-50 backbone trained on real images from LSUN and COCO. The key difference lies in how training fake images are generated: Corvi uses LDM with text prompts corresponding to real image content, while Rajan generates fake images by passing real images through the VAE encoder decoder of LDM. The latter ensures pixel-level alignment between real and fake pairs, isolating decoder artifacts as the sole distinguishing factor. We focus on the final convolutional layer before global average pooling, which contains 2048 channels and produces the feature representation that the linear classifier uses to distinguish real from AI-generated images.

For each detector, we compute RAD values across approximately 6400 inpainted images, averaging per channel scores to obtain a stable ranking. Figure~\ref{fig:feature_align} visualizes activation patterns from the Rajan detector at different points along the RAD distribution. Channels with high RAD values exhibit activation maps that closely match the inpainted mask, indicating selective response to generated content. Channels with strongly negative RAD values show the inverse pattern, activating primarily on real regions.

\paragraph{Alignment Predicts Robustness.}

A natural question arises: do channels that align strongly with generated regions also exhibit greater robustness to post-processing perturbations? To investigate this, we measure the sensitivity of each channel to WEBP compression, a particularly relevant perturbation for these detectors. The LSUN images used to train these detectors were originally compressed using WEBP before being saved in PNG format, while the synthetic training images lacked such compression artifacts. This dataset imbalance caused the detectors to spuriously associate WEBP compression patterns with real images~\cite{rajan2025stay}, a phenomenon analogous to the JPEG bias observed by~\citet{grommelt2024fake}. We quantify channel robustness by computing the mean squared error (MSE) between feature activations before and after applying WEBP compression. Channels whose activations change substantially under this perturbation are deemed less robust.

\begin{figure}[t]
    \centering
    \includegraphics[width=1.0\linewidth]{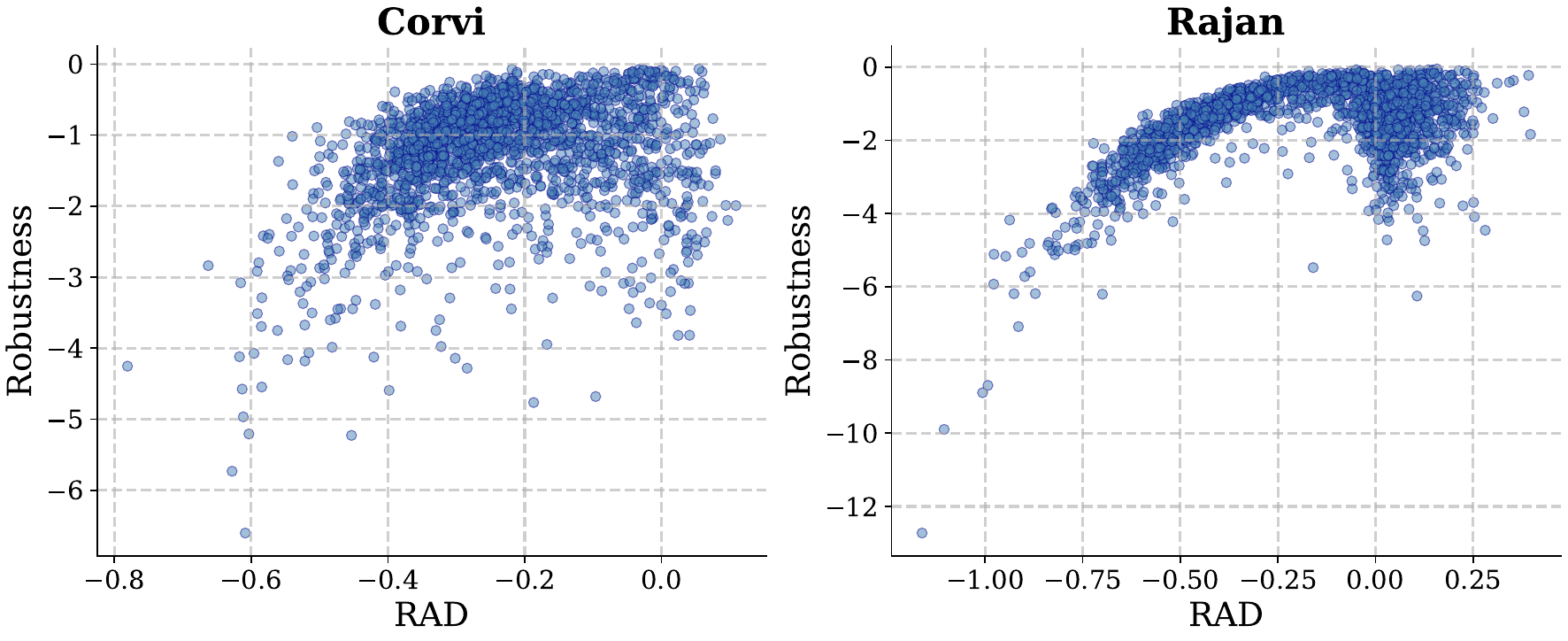}
    \caption{\textbf{Relationship between Feature Alignment and Robustness.} Each point represents one of 2048 channels from the final convolutional layer. The horizontal axis shows RAD (alignment with inpainted regions) and the vertical axis shows robustness measured as negative MSE under WEBP compression. Left: Corvi detector. Right: Rajan detector. Channels at the extremes of the RAD distribution exhibit lower robustness, while channels away from these extremes tend to be more robust.}
    \label{fig:robustness_analysis}
    \vspace{-7px}
\end{figure}


Figure~\ref{fig:robustness_analysis} illustrates the relationship between RAD and robustness, revealing that channels at \emph{both extremes} of the RAD distribution are more susceptible to degradation than those in the middle range. We attribute this trend to the spurious nature of features at these extremes: strongly negative values correspond to dataset-specific signatures like compression artifacts~\cite{rajan2025stay, grommelt2024fake}, while highly positive values overfit to fragile generator shortcuts~\cite{yan2025effort, kashiani2025freqdebias, ma2025specificity}. As both types of signals are easily disrupted by post-processing, their removal enhances stability. In contrast, channels with intermediate RAD values demonstrate greater robustness under perturbation, as they are less reliant on these non-intrinsic artifacts. This suggests that pruning extreme components is an effective strategy to suppress spurious dependencies and enhance overall detector robustness.

\subsection{Feature Pruning and Classifier Refinement}
\label{sec:pruning}

Leveraging the insights from our dissection analysis, we introduce the \emph{Dissect and Prune} strategy. Our analysis indicates that features with extreme alignment scores, whether strongly fake-aligned or strongly real-aligned, are structurally fragile. Therefore, we implement a bilateral pruning mechanism to reduce these spurious dependencies.

\paragraph{Bilateral Pruning Strategy.}
Our analysis (Figure~\ref{fig:robustness_analysis}) indicates that channels at \emph{both} extremes of the RAD distribution are susceptible to post-processing degradation. To address this, we implement a \emph{bilateral pruning} strategy that removes these non-robust components.

Let $\{S_k\}_{k=1}^{D}$ denote the RAD values computed over the diagnostic dataset. We define a binary pruning mask $\mathbf{m} \in \{0, 1\}^D$ to retain only channels within a robust intermediate range. Given lower and upper percentiles $\alpha_{\text{low}}$ and $\alpha_{\text{high}}$, we compute thresholds $\tau_{\text{low}}$ and $\tau_{\text{high}}$ based on the empirical distribution of $S$. The mask for the $k$-th channel is defined as:
\begin{equation}
    m_k = \mathbbm{1}[\tau_{\text{low}} \leq S_k \leq \tau_{\text{high}}],
    \label{eq:mask}
\end{equation}
where $\mathbbm{1}[\cdot]$ is the indicator function. This effectively prunes features that are spuriously correlated with specific compression artifacts (negative extreme) or generator-specific noise (positive extreme), retaining only those stable under perturbation.

\paragraph{Classifier Refinement on Pruned Features.}
We apply this mask to the pre-trained backbone. Since DEAR performs feature selection on the pretrained backbone's representation, we keep the backbone parameters $\theta$ frozen and retrain only the linear classifier on the resulting pruned features. The pruning is applied via element-wise multiplication to the feature tensor $\mathbf{F} \in \mathbb{R}^{D \times h \times w}$ before the global pooling layer:
\begin{equation}
    \tilde{\mathbf{F}} = \mathbf{m} \odot \mathbf{F}.
\end{equation}

With the robust feature subspace defined by $\tilde{\mathbf{F}}$, we reinitialize the weights of the final linear classifier $h_\phi$. Crucially, we fine-tune the classifier using a combined dataset comprising both the original training data $\mathcal{D}_{\text{train}}$ and the diagnostic inpaint data $\mathcal{D}_{\text{inpaint}}$. This joint optimization ensures that the classifier adapts to the pruned feature space while learning to distinguish intrinsic generative artifacts from both global images and localized inpainting regions. The complete procedure is formalized in Algorithm~\ref{alg:dear} in Appendix~\ref{app:impl_detail}.

\section{Experiments}
\label{sec:experiments}

In this section, we empirically validate the effectiveness of DEAR. We investigate \textbf{(1)} whether it maintains robustness against post-processing artifacts such as compression and resizing, \textbf{(2)} whether DEAR generalizes to diverse unseen generative models, and \textbf{(3)} whether it effectively alleviates the prediction asymmetry problem. Further implementation details and analyses are provided in Appendices~\ref{app:impl_detail} and~\ref{app:add_results}.

\subsection{Experimental Setup}

\paragraph{Datasets.}
For real images, we use 3,000 images from the RedCaps dataset~\citep{desai2021redcaps}, which contains diverse web-crawled images with natural variations in content, resolution, and compression. To evaluate generalization across generators, we adopt the test set released by~\citet{rajan2025aligned}, which spans multiple architectures and generation paradigms. Specifically, it includes images from diffusion-based models: Stable Diffusion (SD, 3,000 images)~\citep{rombach2022high}, FLUX (3,000)~\citep{flux}, Kandinsky (3,100)~\citep{razzhigaev2023kandinsky}, Playground (3,150)~\citep{li2024playground}, and PixArt (3,150)~\citep{chen2024pixartalpha}; latent consistency models: LCM (3,146)~\citep{luo2023latent}; alternative architectures: Wuerstchen (3,150)~\citep{pernias2024wuerstchen} and aMUSEd (3,150)~\citep{patil2024amused}; and commercial systems: Midjourney (3,000)~\citep{midjourney}.

To assess performance under realistic deployment conditions, we additionally evaluate on three in the wild benchmarks: Chameleon (6,934 images)~\citep{yan2024sanity}, a curated collection of challenging AI-generated images from online art communities; WildRF (1,051)~\citep{cavia2024real}, images collected from social media platforms using common hashtags; and LOKI (1,921)~\citep{ye2025loki}, a comprehensive benchmark spanning diverse generation sources and specialized domains. These benchmarks are particularly valuable as they contain images that have undergone unknown post-processing operations during upload and sharing. Further details are provided in Appendix~\ref{app:setup_data}.

\definecolor{paleblue}{RGB}{220, 240, 255}
\begin{table*}[ht!]
    \caption{\textbf{Comparison with SoTA detectors.} Methods are grouped by architecture type: CLIP-based, ViT-based (V), Training-free (T-Free), and CNN-based. AUC, R.Acc (real accuracy), and F.Acc (fake accuracy) for each generator. Top: original test images, bottom: post-processed. \textbf{Bold} indicates best, \underline{underline} indicates second best. For training-free methods, we report only threshold-independent AUC (`-') since they lack a calibrated classification threshold.}
    \label{table:main_results}
    \vspace{-0.1cm}
    \centering
    \renewcommand\arraystretch{1.15}
    \setlength{\tabcolsep}{2.5pt}
    \resizebox{1.0\linewidth}{!}{
        \begin{tabular}{clccc|ccc|ccc|ccc|ccc|ccc|ccc|ccc|ccc||ccc}
        \toprule
        & \multicolumn{1}{c}{\bf Original} & \multicolumn{3}{c|}{\textbf{SD}} & \multicolumn{3}{c|}{\textbf{MJ}} & \multicolumn{3}{c|}{\textbf{KD}} & \multicolumn{3}{c|}{\textbf{PG}} & \multicolumn{3}{c|}{\textbf{PixArt}} & \multicolumn{3}{c|}{\textbf{LCM}} & \multicolumn{3}{c|}{\textbf{FLUX}} & \multicolumn{3}{c|}{\textbf{Wuerst.}} & \multicolumn{3}{c||}{\textbf{aMUSEd}} & \multicolumn{3}{c}{\textbf{Avg}} \\
        \cmidrule{3-32}
        & \bf Detectors & AUC & R.Acc & F.Acc & AUC & R.Acc & F.Acc & AUC & R.Acc & F.Acc & AUC & R.Acc & F.Acc & AUC & R.Acc & F.Acc & AUC & R.Acc & F.Acc & AUC & R.Acc & F.Acc & AUC & R.Acc & F.Acc & AUC & R.Acc & F.Acc & AUC & R.Acc & F.Acc \\
        \midrule
        \multirow{4}{*}{\rotatebox{90}{\textit{CLIP}}} & UFD & 66.7 & 95.1 & 17.9 & 54.8 & 95.1 & 13.0 & 70.4 & 95.1 & 23.9 & 72.5 & 95.1 & 20.4 & 73.3 & 95.1 & 20.2 & 72.1 & 95.1 & 23.4 & 21.5 & 95.1 & 0.1 & 93.8 & 95.1 & 74.5 & 95.3 & 95.1 & 77.5 & 68.9 & 95.1 & 30.1 \\
         & C2P-CLIP & 75.3 & 93.0 & 40.9 & 70.8 & 93.0 & 18.7 & 77.9 & 93.0 & 30.6 & 68.3 & 93.0 & 7.3 & 59.6 & 93.0 & 2.2 & 70.5 & 93.0 & 17.6 & 49.9 & 93.0 & 8.0 & 87.1 & 93.0 & 52.0 & 98.8 & 93.0 & 98.3 & 73.1 & 93.0 & 30.6 \\
         & RINE & 97.2 & 92.3 & 90.0 & 90.8 & 92.3 & 71.5 & 98.0 & 92.3 & 92.8 & 85.6 & 92.3 & 60.5 & 96.1 & 92.3 & 87.0 & 99.7 & 92.3 & 99.4 & 69.3 & 92.3 & 30.2 & \underline{99.9} & 92.3 & 99.8 & \underline{99.9} & 92.3 & \textbf{100.0} & 93.0 & 92.3 & 81.2 \\
         & ClipDet & 78.1 & 85.3 & 47.4 & 79.3 & 85.3 & 50.9 & 79.2 & 85.3 & 50.8 & 82.1 & 85.3 & 53.0 & 84.6 & 85.3 & 59.8 & 71.0 & 85.3 & 34.7 & 84.9 & 85.3 & 64.9 & 93.0 & 85.3 & 87.0 & 91.2 & 85.3 & 80.5 & 82.6 & 85.3 & 58.8 \\
        \cmidrule{1-32}
        \multirow{1}{*}{\rotatebox{90}{\textit{V}}} & CoDE & 87.4 & 85.5 & 71.9 & 81.0 & 85.5 & 60.1 & 68.9 & 85.5 & 38.6 & 86.8 & 85.5 & 70.5 & 93.2 & 85.5 & 86.5 & 85.9 & 85.5 & 69.2 & 62.3 & 85.5 & 25.3 & 70.3 & 85.5 & 42.0 & 74.9 & 85.5 & 46.3 & 79.0 & 85.5 & 56.7 \\
        \cmidrule{1-32}
        \multirow{2}{*}{\rotatebox{90}{\textit{T-Free}}} & AEROBLADE & 98.9 & - & - & 99.4 & - & - & 98.4 & - & - & 70.9 & - & - & 94.3 & - & - & 99.0 & - & - & 78.3 & - & - & 81.9 & - & - & 93.0 & - & - & 90.5 & - & - \\
         & WaRPAD & 77.2 & - & - & 74.9 & - & - & 77.0 & - & - & 73.2 & - & - & 66.1 & - & - & 66.1 & - & - & 56.1 & - & - & 92.9 & - & - & 96.2 & - & - & 75.5 & - & - \\
        \cmidrule{1-32}
        \multirow{12}{*}{\rotatebox{90}{\textit{CNN}}} & NPR & 80.5 & 67.2 & 80.4 & 93.9 & 67.2 & 94.2 & 91.6 & 67.2 & 94.5 & 95.8 & 67.2 & 99.4 & 96.1 & 67.2 & \underline{99.9} & 93.5 & 67.2 & 98.4 & 97.5 & 67.2 & \underline{99.7} & 98.2 & 67.2 & 99.1 & 95.9 & 67.2 & 97.3 & 93.7 & 67.2 & 95.9 \\
         & SAFE & 83.5 & 97.8 & 65.0 & 99.0 & 97.8 & 92.2 & 99.6 & 97.8 & 97.9 & 98.7 & 97.8 & 89.3 & 99.5 & 97.8 & 97.7 & 99.4 & 97.8 & 96.6 & 99.6 & 97.8 & 98.2 & 99.7 & 97.8 & \textbf{100.0} & 99.6 & 97.8 & 98.7 & 97.6 & 97.8 & 92.8 \\
         & AIDE & 91.4 & 88.8 & 78.7 & 79.3 & 88.8 & 55.5 & 95.1 & 88.8 & 87.5 & 95.0 & 88.8 & 87.0 & 98.1 & 88.8 & 96.4 & 91.5 & 88.8 & 77.5 & 96.1 & 88.8 & 90.3 & 94.4 & 88.8 & 87.5 & 94.0 & 88.8 & 82.7 & 92.8 & 88.8 & 82.6 \\
         & FerretNet & 90.8 & 95.8 & 73.0 & 99.2 & 95.8 & 97.1 & 99.4 & 95.8 & 97.5 & \underline{99.9} & 95.8 & \underline{99.8} & \underline{99.9} & 95.8 & \underline{99.9} & \underline{99.9} & 95.8 & \textbf{100.0} & \underline{99.8} & 95.8 & 98.9 & 99.7 & 95.8 & 99.2 & 99.8 & 95.8 & 99.1 & 98.7 & 95.8 & 96.0 \\
         & LaDeDa & 90.9 & 97.8 & 71.7 & 97.7 & 97.8 & 83.7 & 98.4 & 97.8 & 85.9 & 97.4 & 97.8 & 71.1 & 97.5 & 97.8 & 66.4 & 99.0 & 97.8 & 88.6 & 99.7 & 97.8 & 97.4 & 99.8 & 97.8 & 99.2 & 99.8 & 97.8 & 98.4 & 97.8 & 97.8 & 84.7 \\
         & DRCT & 98.6 & 96.0 & 94.7 & 96.8 & 96.0 & 84.8 & \underline{99.8} & 96.0 & 99.1 & 90.6 & 96.0 & 64.4 & 75.5 & 96.0 & 46.3 & 96.8 & 96.0 & 84.8 & 82.9 & 96.0 & 49.2 & 97.0 & 96.0 & 79.6 & 94.1 & 96.0 & 65.0 & 92.4 & 96.0 & 74.2 \\
         & Corvi & \textbf{100.0} & \textbf{99.9} & 99.5 & \underline{99.9} & \textbf{99.9} & 97.7 & \textbf{100.0} & \textbf{99.9} & \textbf{100.0} & \underline{99.9} & \textbf{99.9} & 91.3 & \textbf{100.0} & \textbf{99.9} & \textbf{100.0} & \textbf{100.0} & \textbf{99.9} & 99.7 & 87.8 & \textbf{99.9} & 21.5 & \textbf{100.0} & \textbf{99.9} & \textbf{100.0} & 98.9 & \textbf{99.9} & 68.5 & 98.5 & \textbf{99.9} & 86.5 \\
         & Corvi+ & \textbf{100.0} & \underline{99.1} & \underline{99.9} & \textbf{100.0} & \underline{99.1} & 99.2 & \textbf{100.0} & \underline{99.1} & \textbf{100.0} & \textbf{100.0} & \underline{99.1} & \textbf{100.0} & \textbf{100.0} & \underline{99.1} & \textbf{100.0} & \textbf{100.0} & \underline{99.1} & \textbf{100.0} & \textbf{100.0} & \underline{99.1} & 99.6 & \textbf{100.0} & \underline{99.1} & \textbf{100.0} & \textbf{100.0} & \underline{99.1} & \underline{99.9} & \textbf{100.0} & \underline{99.1} & \underline{99.8} \\
        \rowcolor{paleblue}  & \textbf{DEAR-c} & \textbf{100.0} & 96.3 & \underline{99.9} & \textbf{100.0} & 96.3 & \textbf{100.0} & \textbf{100.0} & 96.3 & \textbf{100.0} & \textbf{100.0} & 96.3 & \textbf{100.0} & \textbf{100.0} & 96.3 & \textbf{100.0} & \textbf{100.0} & 96.3 & \textbf{100.0} & \textbf{100.0} & 96.3 & \textbf{100.0} & \textbf{100.0} & 96.3 & \textbf{100.0} & \textbf{100.0} & 96.3 & \underline{99.9} & \textbf{100.0} & 96.3 & \textbf{100.0} \\
         & Rajan & \underline{99.9} & \textbf{99.9} & 99.2 & \underline{99.9} & \textbf{99.9} & 98.0 & \textbf{100.0} & \textbf{99.9} & \underline{99.9} & \textbf{100.0} & \textbf{99.9} & 95.4 & \textbf{100.0} & \textbf{99.9} & \textbf{100.0} & \textbf{100.0} & \textbf{99.9} & \underline{99.8} & 89.2 & \textbf{99.9} & 12.0 & 98.3 & \textbf{99.9} & 45.7 & 83.9 & \textbf{99.9} & 24.8 & 96.8 & \textbf{99.9} & 75.0 \\
         & Rajan+ & \textbf{100.0} & \textbf{99.9} & \textbf{100.0} & \textbf{100.0} & \textbf{99.9} & 99.0 & \textbf{100.0} & \textbf{99.9} & \textbf{100.0} & \textbf{100.0} & \textbf{99.9} & \textbf{100.0} & \textbf{100.0} & \textbf{99.9} & \textbf{100.0} & \textbf{100.0} & \textbf{99.9} & \textbf{100.0} & 97.4 & \textbf{99.9} & 29.6 & 99.2 & \textbf{99.9} & 71.0 & \underline{99.9} & \textbf{99.9} & 93.9 & 99.6 & \textbf{99.9} & 88.2 \\
        \rowcolor{paleblue}  & \textbf{DEAR-r} & \underline{99.9} & 97.4 & 99.8 & \underline{99.9} & 97.4 & \underline{99.8} & \textbf{100.0} & 97.4 & \textbf{100.0} & \textbf{100.0} & 97.4 & \textbf{100.0} & \textbf{100.0} & 97.4 & \textbf{100.0} & \textbf{100.0} & 97.4 & \textbf{100.0} & 99.5 & 97.4 & 96.6 & \textbf{100.0} & 97.4 & \underline{99.9} & 99.5 & 97.4 & 96.6 & \underline{99.9} & 97.4 & 99.2 \\
        \midrule
        \midrule
        & \multicolumn{1}{c}{\bf Post-proc.} & \multicolumn{3}{c|}{\textbf{SD}} & \multicolumn{3}{c|}{\textbf{MJ}} & \multicolumn{3}{c|}{\textbf{KD}} & \multicolumn{3}{c|}{\textbf{PG}} & \multicolumn{3}{c|}{\textbf{PixArt}} & \multicolumn{3}{c|}{\textbf{LCM}} & \multicolumn{3}{c|}{\textbf{FLUX}} & \multicolumn{3}{c|}{\textbf{Wuerst.}} & \multicolumn{3}{c||}{\textbf{aMUSEd}} & \multicolumn{3}{c}{\textbf{Avg}} \\
        \cmidrule{3-32}
        & \bf Detectors & AUC & R.Acc & F.Acc & AUC & R.Acc & F.Acc & AUC & R.Acc & F.Acc & AUC & R.Acc & F.Acc & AUC & R.Acc & F.Acc & AUC & R.Acc & F.Acc & AUC & R.Acc & F.Acc & AUC & R.Acc & F.Acc & AUC & R.Acc & F.Acc & AUC & R.Acc & F.Acc \\
        \midrule
        \multirow{4}{*}{\rotatebox{90}{\textit{CLIP}}} & UFD & 59.2 & 93.2 & 13.1 & 42.9 & 93.2 & 8.2 & 56.4 & 93.2 & 14.5 & 53.3 & 93.2 & 7.5 & 54.5 & 93.2 & 9.5 & 57.2 & 93.2 & 12.9 & 25.0 & 93.2 & 0.6 & 73.7 & 93.2 & 33.2 & 85.3 & 93.2 & 50.7 & 56.4 & 93.2 & 16.7 \\
         & C2P-CLIP & 60.9 & 96.6 & 14.3 & 68.5 & 96.6 & 9.7 & 70.6 & 96.6 & 14.0 & 74.6 & 96.6 & 9.6 & 70.6 & 96.6 & 3.2 & 61.0 & 96.6 & 6.2 & 35.5 & 96.6 & 0.6 & 71.7 & 96.6 & 21.1 & 93.0 & 96.6 & 62.5 & 67.4 & 96.6 & 15.7 \\
         & RINE & 76.0 & 77.5 & 61.7 & 53.3 & 77.5 & 28.5 & 64.8 & 77.5 & 41.9 & 59.7 & 77.5 & 37.6 & 45.3 & 77.5 & 20.0 & 70.3 & 77.5 & 48.5 & 45.7 & 77.5 & 18.2 & 85.2 & 77.5 & 78.6 & 86.4 & 77.5 & 80.8 & 65.2 & 77.5 & 46.2 \\
         & ClipDet & 71.9 & 76.6 & 52.8 & 75.2 & 76.6 & 59.4 & 75.4 & 76.6 & 59.6 & 76.9 & 76.6 & 60.6 & 79.3 & 76.6 & 66.2 & 76.7 & 76.6 & 61.2 & 79.8 & 76.6 & 66.7 & 89.2 & 76.6 & \underline{85.2} & 87.7 & 76.6 & 82.3 & 79.1 & 76.6 & 66.0 \\
        \cmidrule{1-32}
        \multirow{1}{*}{\rotatebox{90}{\textit{V}}} & CoDE & 82.2 & 81.0 & 68.4 & 68.8 & 81.0 & 47.6 & 64.4 & 81.0 & 41.5 & 69.4 & 81.0 & 46.3 & 76.2 & 81.0 & 59.6 & 67.4 & 81.0 & 44.1 & 63.6 & 81.0 & 38.4 & 70.3 & 81.0 & 49.0 & 70.2 & 81.0 & 47.9 & 70.3 & 81.0 & 49.2 \\
        \cmidrule{1-32}
        \multirow{2}{*}{\rotatebox{90}{\textit{T-Free}}} & AEROBLADE & 56.1 & - & - & 78.0 & - & - & 63.8 & - & - & 56.7 & - & - & 58.2 & - & - & 42.5 & - & - & 54.6 & - & - & 56.6 & - & - & 67.0 & - & - & 59.3 & - & - \\
         & WaRPAD & 64.5 & - & - & 59.8 & - & - & 65.3 & - & - & 60.1 & - & - & 57.1 & - & - & 52.4 & - & - & 35.0 & - & - & 82.2 & - & - & 91.2 & - & - & 63.1 & - & - \\
        \cmidrule{1-32}
        \multirow{12}{*}{\rotatebox{90}{\textit{CNN}}} & NPR & 50.3 & 94.2 & 10.0 & 63.0 & 94.2 & 12.6 & 62.7 & 94.2 & 19.8 & 66.1 & 94.2 & 16.3 & 75.6 & 94.2 & 26.2 & 35.9 & 94.2 & 5.2 & 41.9 & 94.2 & 5.6 & 56.0 & 94.2 & 11.6 & 27.9 & 94.2 & 2.1 & 53.3 & 94.2 & 12.2 \\
         & SAFE & 53.9 & 99.5 & 1.4 & 40.1 & 99.5 & 0.7 & 48.1 & 99.5 & 1.4 & 38.1 & 99.5 & 0.3 & 35.9 & 99.5 & 0.7 & 47.3 & 99.5 & 0.0 & 48.2 & 99.5 & 0.5 & 51.7 & 99.5 & 1.0 & 57.5 & 99.5 & 0.6 & 46.8 & 99.5 & 0.7 \\
         & AIDE & 65.8 & 90.3 & 23.4 & 64.9 & 90.3 & 17.2 & 68.3 & 90.3 & 25.5 & 61.1 & 90.3 & 9.9 & 69.8 & 90.3 & 23.2 & 58.3 & 90.3 & 14.8 & 65.8 & 90.3 & 21.3 & 75.6 & 90.3 & 27.7 & 76.1 & 90.3 & 30.5 & 67.3 & 90.3 & 21.5 \\
         & FerretNet & 56.5 & 97.7 & 2.7 & 44.7 & 97.7 & 1.1 & 59.1 & 97.7 & 2.6 & 64.9 & 97.7 & 3.6 & 73.1 & 97.7 & 6.0 & 49.9 & 97.7 & 1.1 & 58.7 & 97.7 & 1.7 & 65.6 & 97.7 & 1.7 & 60.9 & 97.7 & 0.3 & 59.3 & 97.7 & 2.3 \\
         & LaDeDa & 45.4 & \underline{99.8} & 1.5 & 57.1 & \underline{99.8} & 0.2 & 61.6 & \underline{99.8} & 0.9 & 64.6 & \underline{99.8} & 0.0 & 73.8 & \underline{99.8} & 0.1 & 33.6 & \underline{99.8} & 0.0 & 42.1 & \underline{99.8} & 0.1 & 51.9 & \underline{99.8} & 0.2 & 33.9 & \underline{99.8} & 0.0 & 51.6 & \underline{99.8} & 0.3 \\
         & DRCT & 90.5 & 86.2 & 80.6 & 79.5 & 86.2 & 59.4 & 86.7 & 86.2 & 74.2 & 71.7 & 86.2 & 46.6 & 72.1 & 86.2 & 52.2 & 73.3 & 86.2 & 41.4 & 67.3 & 86.2 & 39.0 & 77.1 & 86.2 & 50.6 & 85.6 & 86.2 & 65.7 & 78.2 & 86.2 & 56.6 \\
         & Corvi & 91.1 & 97.3 & 78.4 & 80.0 & 97.3 & 50.2 & 80.7 & 97.3 & 57.0 & 66.8 & 97.3 & 18.5 & 75.7 & 97.3 & 63.4 & 81.1 & 97.3 & 46.0 & 64.2 & 97.3 & 21.1 & 82.8 & 97.3 & 59.3 & 90.7 & 97.3 & 35.8 & 79.2 & 97.3 & 47.7 \\
         & Corvi+ & 95.5 & 96.8 & 72.5 & 85.1 & 96.8 & 42.5 & 89.9 & 96.8 & 57.1 & 88.0 & 96.8 & 44.4 & 93.2 & 96.8 & 72.0 & 91.8 & 96.8 & 53.3 & 84.3 & 96.8 & 28.6 & 91.8 & 96.8 & 55.7 & 85.7 & 96.8 & 33.1 & 89.5 & 96.8 & 51.0 \\
        \rowcolor{paleblue}  & \textbf{DEAR-c} & 96.5 & 76.4 & \textbf{96.2} & 89.0 & 76.4 & \underline{84.9} & 92.4 & 76.4 & \underline{89.4} & 92.9 & 76.4 & \underline{91.0} & 93.8 & 76.4 & \underline{91.0} & 96.1 & 76.4 & \underline{96.5} & 82.4 & 76.4 & \underline{73.6} & \underline{94.8} & 76.4 & \textbf{93.3} & 93.2 & 76.4 & \textbf{96.0} & 92.3 & 76.4 & \textbf{90.2} \\
         & Rajan & \underline{97.7} & \textbf{99.9} & 86.3 & \underline{94.5} & \textbf{99.9} & 69.9 & \underline{92.6} & \textbf{99.9} & 65.7 & \underline{93.8} & \textbf{99.9} & 59.7 & \underline{93.9} & \textbf{99.9} & 84.5 & \underline{99.7} & \textbf{99.9} & 93.0 & 81.2 & \textbf{99.9} & 15.4 & 87.1 & \textbf{99.9} & 15.6 & 95.9 & \textbf{99.9} & 33.8 & \underline{92.9} & \textbf{99.9} & 58.2 \\
         & Rajan+ & 96.1 & 99.1 & 83.4 & 86.1 & 99.1 & 64.4 & 91.6 & 99.1 & 71.6 & 89.3 & 99.1 & 68.7 & 91.4 & 99.1 & 79.9 & 97.3 & 99.1 & 84.8 & \underline{84.7} & 99.1 & 27.8 & 82.1 & 99.1 & 17.4 & \underline{96.6} & 99.1 & 63.0 & 90.6 & 99.1 & 62.3 \\
        \rowcolor{paleblue}  & \textbf{DEAR-r} & \textbf{98.6} & 95.1 & \underline{95.7} & \textbf{95.3} & 95.1 & \textbf{86.0} & \textbf{97.3} & 95.1 & \textbf{91.4} & \textbf{97.9} & 95.1 & \textbf{92.4} & \textbf{97.6} & 95.1 & \textbf{93.2} & \textbf{99.8} & 95.1 & \textbf{99.0} & \textbf{94.4} & 95.1 & \textbf{77.6} & \textbf{95.7} & 95.1 & 82.7 & \textbf{97.6} & 95.1 & \underline{87.8} & \textbf{97.1} & 95.1 & \underline{89.5} \\
        \bottomrule
        \end{tabular}
    }
\end{table*}

\definecolor{paleblue}{RGB}{220, 240, 255}
\begin{table}[ht!]
    \caption{\textbf{Additional results on Chameleon, Loki, and WildRF.} Methods are grouped by architecture type: CLIP-based, ViT-based (V), Training-free (T-Free), and CNN-based. \textbf{Bold} indicates best, \underline{underline} indicates second best. For training-free methods, we report only threshold-independent AUC (`-') since they lack a calibrated classification threshold.}
    \label{table:appendix_results}
    \vspace{-0.1cm}
    \centering
    \renewcommand\arraystretch{1.0}
    \setlength{\tabcolsep}{3pt}
    \small
    \resizebox{1.0\columnwidth}{!}{
    \begin{tabular}{clccc|ccc|ccc||ccc}
    \toprule
        & \multicolumn{1}{c}{\bf Original} & \multicolumn{3}{c|}{\textbf{Chameleon}} & \multicolumn{3}{c|}{\textbf{Loki}} & \multicolumn{3}{c||}{\textbf{WildRF}} & \multicolumn{3}{c}{\textbf{Avg}} \\
        \cmidrule{3-14}
        & \bf Detectors & AUC & R.Acc & F.Acc & AUC & R.Acc & F.Acc & AUC & R.Acc & F.Acc & AUC & R.Acc & F.Acc \\
        \midrule
        \multirow{4}{*}{\rotatebox{90}{\textit{CLIP}}} & UFD & 55.4 & 97.8 & 3.6 & 67.5 & 95.1 & 27.9 & 59.1 & 77.1 & 38.0 & 60.7 & 90.0 & 23.2 \\
         & C2P-CLIP & 44.2 & 99.5 & 2.8 & 65.8 & 93.0 & 27.9 & 74.9 & 90.6 & 36.6 & 61.6 & 94.3 & 22.4 \\
         & RINE & 39.2 & 60.7 & 23.5 & 76.8 & 92.3 & 47.0 & 78.2 & 94.1 & 45.8 & 64.7 & 82.4 & 38.8 \\
         & ClipDet & 52.7 & 74.1 & 28.9 & 69.2 & 85.3 & 37.6 & \textbf{89.4} & 93.9 & 56.3 & 70.4 & 84.5 & 40.9 \\
        \cmidrule{1-14}
        \multirow{1}{*}{\rotatebox{90}{\textit{V}}} & CoDE & 81.6 & 78.3 & \underline{69.6} & 61.9 & 85.5 & 39.8 & 72.5 & 79.5 & 56.3 & 72.0 & 81.1 & 55.2 \\
        \cmidrule{1-14}
        \multirow{2}{*}{\rotatebox{90}{\textit{T-Free}}} & AEROBLADE & 74.8 & - & - & 63.5 & - & - & 72.1 & - & - & 70.1 & - & - \\
         & WaRPAD & 55.2 & - & - & 67.5 & - & - & 79.8 & - & - & 67.5 & - & - \\
        \cmidrule{1-14}
        \multirow{12}{*}{\rotatebox{90}{\textit{CNN}}} & NPR & 52.1 & 98.8 & 8.0 & 69.3 & 67.2 & \textbf{56.4} & 69.9 & 65.1 & 61.3 & 63.8 & 77.0 & 41.9 \\
         & SAFE & 57.1 & \underline{99.6} & 5.5 & 69.0 & 97.8 & 47.4 & 69.0 & 98.8 & 31.3 & 65.1 & 98.7 & 28.1 \\
         & AIDE & 73.7 & 93.0 & 25.7 & 68.9 & 88.8 & 45.3 & 72.9 & 89.4 & 45.1 & 71.8 & 90.4 & 38.7 \\
         & FerretNet & 42.9 & 99.2 & 5.8 & 68.5 & 95.8 & 46.9 & 74.4 & 98.5 & 32.7 & 61.9 & 97.9 & 28.4 \\
         & LaDeDa & 50.9 & \textbf{100.0} & 5.3 & 63.9 & 97.8 & 45.5 & 78.6 & 98.4 & 34.2 & 64.5 & 98.7 & 28.3 \\
         & DRCT & 74.6 & 79.9 & 57.9 & \textbf{85.9} & 96.0 & 50.6 & 76.7 & 82.7 & 59.1 & 79.1 & 86.2 & 55.9 \\
         & Corvi & 73.2 & 98.9 & 19.4 & 72.8 & \textbf{99.9} & 40.9 & 85.2 & \underline{99.4} & 56.8 & 77.1 & 99.4 & 39.0 \\
         & Corvi+ & \underline{83.6} & 99.5 & 18.8 & 78.2 & \underline{99.1} & 47.1 & 82.4 & 99.0 & 56.5 & 81.4 & 99.2 & 40.8 \\
        \rowcolor{paleblue}  & \textbf{DEAR-c} & \textbf{85.6} & 76.2 & \textbf{77.9} & 75.4 & 96.3 & \underline{53.2} & \underline{87.8} & 94.6 & \textbf{70.8} & \textbf{82.9} & 89.0 & \textbf{67.3} \\
         & Rajan & 79.9 & 98.9 & 26.8 & 74.6 & \textbf{99.9} & 35.6 & 85.7 & \textbf{99.8} & 57.6 & 80.1 & \textbf{99.6} & 40.0 \\
         & Rajan+ & 75.4 & 99.3 & 30.1 & 73.9 & \textbf{99.9} & 38.3 & 83.8 & \underline{99.4} & 62.9 & 77.7 & \underline{99.5} & 43.8 \\
        \rowcolor{paleblue}  & \textbf{DEAR-r} & 78.9 & 81.7 & 64.2 & \underline{78.5} & 97.4 & 49.3 & 87.7 & 97.0 & \underline{68.0} & \underline{81.7} & 92.1 & \underline{60.5} \\
        \midrule
        \midrule
        & \multicolumn{1}{c}{\bf Post-proc.} & \multicolumn{3}{c|}{\textbf{Chameleon}} & \multicolumn{3}{c|}{\textbf{Loki}} & \multicolumn{3}{c||}{\textbf{WildRF}} & \multicolumn{3}{c}{\textbf{Avg}} \\
        \cmidrule{3-14}
        & \bf Detectors & AUC & R.Acc & F.Acc & AUC & R.Acc & F.Acc & AUC & R.Acc & F.Acc & AUC & R.Acc & F.Acc \\
        \midrule
        \multirow{4}{*}{\rotatebox{90}{\textit{CLIP}}} & UFD & 44.5 & 96.0 & 2.4 & 58.0 & 93.2 & 19.1 & 60.2 & 92.0 & 21.6 & 54.3 & 93.8 & 14.4 \\
         & C2P-CLIP & 52.9 & \underline{99.8} & 1.0 & 56.9 & 96.6 & 8.4 & 73.9 & 94.1 & 21.9 & 61.2 & 96.8 & 10.4 \\
         & RINE & 35.5 & 63.1 & 18.1 & 58.2 & 77.5 & 36.2 & 75.5 & 91.6 & 35.1 & 56.4 & 77.4 & 29.8 \\
         & ClipDet & 45.2 & 45.0 & \underline{48.1} & 62.7 & 76.6 & 39.8 & \underline{86.7} & 87.9 & 65.2 & 64.9 & 69.8 & 51.0 \\
        \cmidrule{1-14}
        \multirow{1}{*}{\rotatebox{90}{\textit{V}}} & CoDE & 70.5 & 83.2 & 44.1 & 58.5 & 81.0 & 35.6 & 72.7 & 83.8 & 51.0 & 67.2 & 82.7 & 43.5 \\
        \cmidrule{1-14}
        \multirow{2}{*}{\rotatebox{90}{\textit{T-Free}}} & AEROBLADE & 60.3 & - & - & 53.6 & - & - & 58.3 & - & - & 57.4 & - & - \\
         & WaRPAD & 47.8 & - & - & 59.0 & - & - & 74.1 & - & - & 60.3 & - & - \\
        \cmidrule{1-14}
        \multirow{12}{*}{\rotatebox{90}{\textit{CNN}}} & NPR & 47.6 & 98.6 & 2.1 & 57.6 & 94.2 & 8.1 & 52.2 & 90.9 & 14.5 & 52.5 & 94.6 & 8.2 \\
         & SAFE & 53.4 & 99.5 & 0.1 & 37.9 & 99.5 & 0.4 & 57.4 & 99.0 & 1.2 & 49.5 & 99.3 & 0.6 \\
         & AIDE & 64.9 & 90.8 & 19.1 & 51.5 & 90.3 & 11.9 & 67.3 & 94.9 & 21.2 & 61.2 & 92.0 & 17.4 \\
         & FerretNet & 41.2 & 99.1 & 0.5 & 42.7 & 97.7 & 1.3 & 59.2 & 99.3 & 2.5 & 47.7 & 98.7 & 1.4 \\
         & LaDeDa & 51.7 & \textbf{99.9} & 0.0 & 44.8 & \underline{99.8} & 1.1 & 54.8 & \textbf{99.8} & 0.6 & 50.4 & \textbf{99.9} & 0.6 \\
         & DRCT & 66.1 & 78.5 & 44.1 & 70.9 & 86.2 & 44.4 & 78.0 & 86.6 & 53.2 & 71.7 & 83.7 & 47.2 \\
         & Corvi & 57.1 & 93.0 & 13.6 & 66.9 & 97.3 & 34.4 & 84.1 & 99.4 & 34.3 & 69.4 & 96.6 & 27.5 \\
         & Corvi+ & \underline{72.9} & 96.0 & 10.5 & 70.7 & 96.8 & 32.2 & 78.6 & \underline{99.7} & 30.7 & 74.1 & 97.5 & 24.4 \\
        \rowcolor{paleblue}  & \textbf{DEAR-c} & 70.7 & 72.1 & \textbf{58.5} & 72.4 & 76.4 & \textbf{59.6} & \textbf{89.6} & 94.2 & \textbf{72.3} & \underline{77.5} & 80.9 & \textbf{63.5} \\
         & Rajan & 66.8 & 99.6 & 9.5 & \underline{73.4} & \textbf{99.9} & 33.5 & 85.6 & \textbf{99.8} & 43.3 & 75.3 & \underline{99.8} & 28.8 \\
         & Rajan+ & 64.6 & 99.0 & 14.5 & 70.8 & 99.1 & 37.3 & 81.4 & \textbf{99.8} & 43.9 & 72.3 & 99.3 & 31.9 \\
        \rowcolor{paleblue}  & \textbf{DEAR-r} & \textbf{73.4} & 93.1 & 42.4 & \textbf{74.7} & 95.1 & \underline{47.9} & 86.3 & 97.0 & \underline{65.6} & \textbf{78.1} & 95.0 & \underline{52.0} \\
        \bottomrule
    \end{tabular}
    }
\end{table}

\paragraph{Baselines.}
We compare against a comprehensive set of state-of-the-art detectors spanning multiple paradigms. From CNN based methods, we include Corvi~\citep{corvi2023detection}, Rajan~\citep{rajan2025aligned}, NPR~\citep{tan2024rethinking}, SAFE~\citep{li2025improving}, AIDE~\citep{yan2024sanity}, FerretNet~\citep{liang2025ferretnet}, LaDeDa~\citep{cavia2024real}, and DRCT~\citep{chen2024drct}. From CLIP based approaches, we evaluate UFD~\citep{ojha2023towards}, C2P-CLIP~\citep{tan2025c2p}, RINE~\citep{koutlis2024leveraging}, and ClipDet~\citep{cozzolino2024raising}. We also consider the Vision Transformer based method CoDE~\citep{baraldi2024contrasting}, as well as training-free detectors AEROBLADE~\citep{ricker2024aeroblade} and WaRPAD~\citep{choi2025training}. Additionally, we compare against Corvi+ and Rajan+~\citep{rajan2025stay}, which apply the Stay-Positive algorithm to retrain only the last layer with non-negative weight constraints. Detailed descriptions of each baseline are provided in Appendix~\ref{app:setup_baseline}.

We apply DEAR to two representative detectors, Corvi and Rajan, and refer to the resulting methods as \textbf{DEAR-c} and \textbf{DEAR-r}, respectively. Both employ a ResNet-50 backbone trained on real images from LSUN and COCO, but differ in how synthetic training data is constructed: Corvi uses LDM-generated images conditioned on text prompts, while Rajan uses VAE reconstructions of real images to ensure pixel-level alignment between training pairs. 

\subsection{Robustness to Post-Processing}
A practical detector must maintain reliable performance when images undergo common transformations during storage and transmission. Following the prior evaluation protocol~\citep{rajan2025stay}, we construct a post-processed test set by randomly applying compression, resizing, and color jittering to the synthetic images. Table~\ref{table:main_results} (bottom) summarizes the results. While most baselines suffer noticeable performance degradation under these perturbations, DEAR-c and DEAR-r exhibit substantially improved robustness. The performance gap between DEAR and base detectors becomes especially pronounced on challenging generators like FLUX, where post-processing often removes the fragile artifacts that conventional detectors rely upon. This observation aligns with the intended behavior of DEAR, where bilateral pruning removes these fragile features and inpaint-augmented classifier refinement focuses the detector on intrinsic generative artifacts that persist through post-processing. Additional ablation studies comparing DEAR against detectors trained from scratch with inpainted data augmentation and analyzing sensitivity to the pruning ratio $\alpha$ are provided in Appendix~\ref{app:add_results} (Figures~\ref{fig:appendix_comparison_avg} and~\ref{fig:appendix_comparison}).

\subsection{Evaluation on the Wild Benchmarks}

To assess robustness under real deployment conditions, we evaluate on benchmarks collected from social media platforms and online art communities. As shown in Table~\ref{table:appendix_results}, DEAR-c and DEAR-r demonstrate strong performance on Chameleon~\citep{yan2024sanity}, WildRF~\citep{cavia2024real}, and LOKI~\citep{ye2025loki}, confirming that the robustness gains observed on controlled post-processed data transfer to in-the-wild scenarios.

\subsection{Alleviating Asymmetry and Improving Calibration}

\begin{figure}[t]
    \centering
    \includegraphics[width=1.0\linewidth]{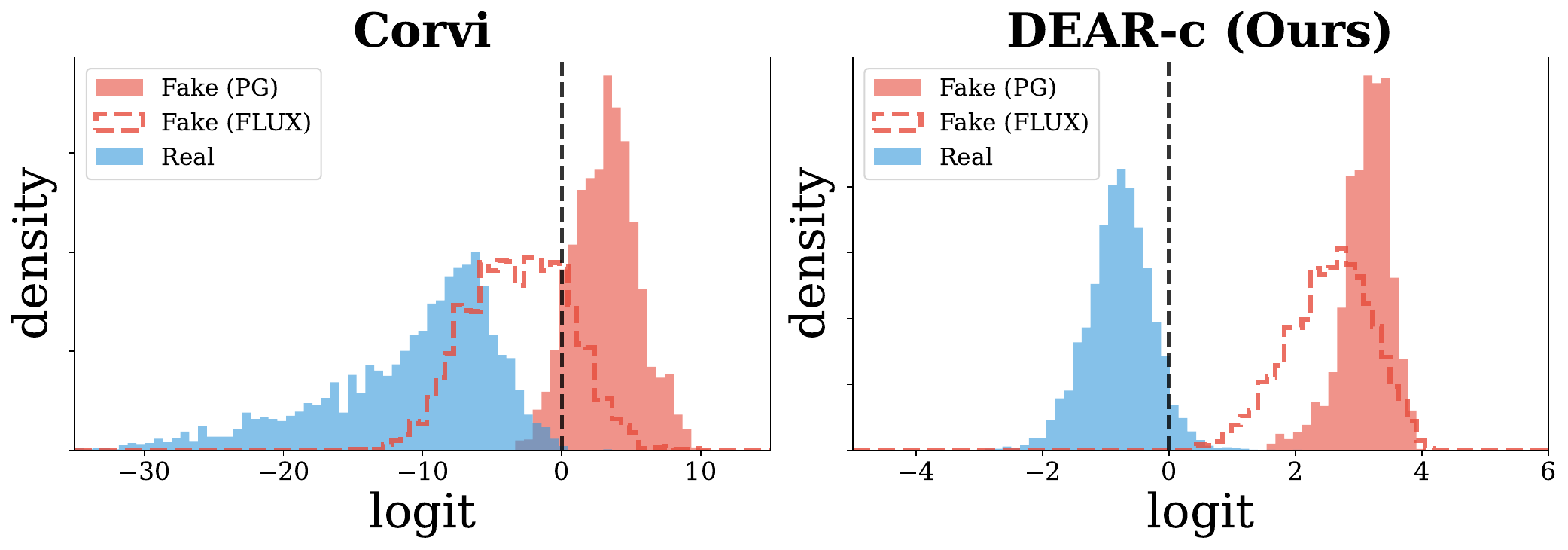}
    \includegraphics[width=1.0\linewidth]{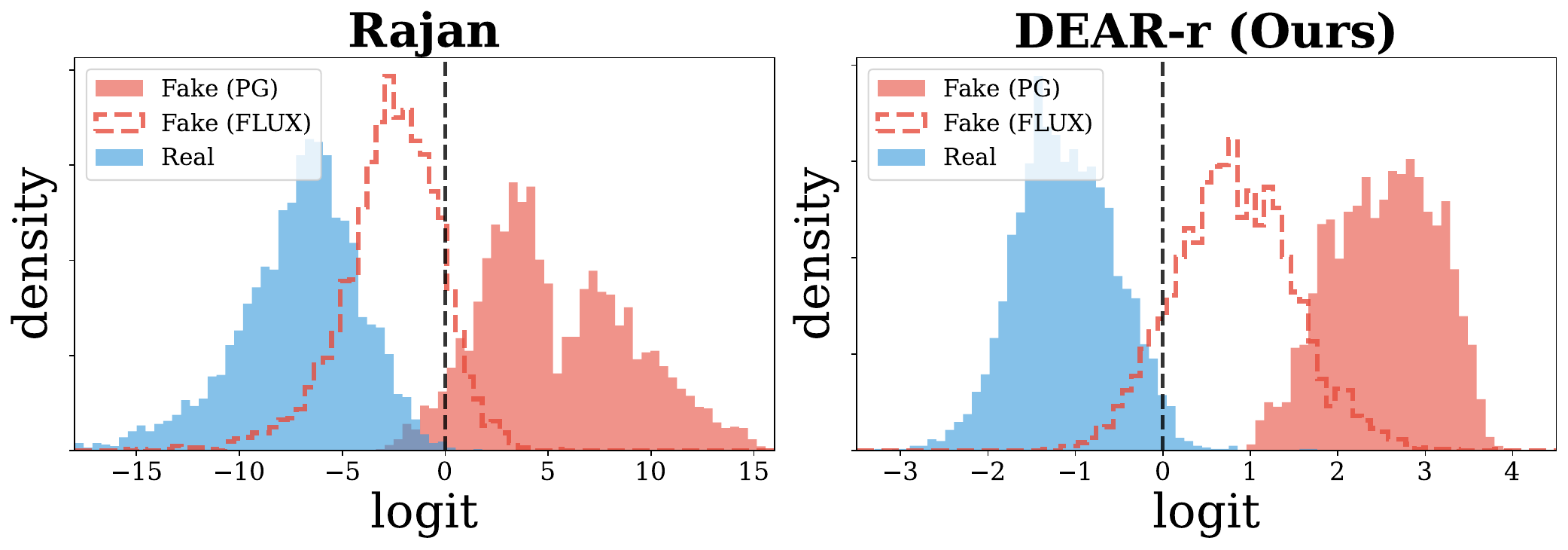}
    \caption{\textbf{Score distribution shift.} Baseline detectors (left) exhibit severe distribution shift on unseen generators, while DEAR variants (right) maintain stable fake score distributions above the decision threshold.}
    \label{fig:score_distribution_shift}
\vspace{-7px}
\end{figure}

Prediction asymmetry is characterized primarily by low F.Acc, complicating the establishment of a universal decision threshold despite high threshold-independent metrics like AUC. For instance, Table~\ref{table:main_results} shows that while DRCT achieves an average AUC of 92.4, its fixed-threshold F.Acc drops to 74.2, whereas R.Acc remains near-ceiling (96.0). This divergence indicates a decision rule heavily biased toward the real class, resulting in a substantial miss rate for fake images under standard thresholds.

In contrast, DEAR enhances generalizability by maintaining stable fake score distributions across diverse generative models, thereby mitigating the need for dataset-specific calibration. Figure~\ref{fig:score_distribution_shift} illustrates this on original test images by comparing score distributions for Playground (4-channel) and FLUX (16-channel). Corvi's fake scores for FLUX shift dramatically below the decision threshold because the detector misinterprets the fine-grained details preserved by FLUX's 16-channel architecture as indicators of authenticity. By pruning these spurious dependencies, DEAR-c aligns score distributions across disparate architectures, ensuring reliable detection with a single threshold.

\section{Related Work}


Early detection methods focused on GAN-generated images, training CNN classifiers on generator-specific traces~\citep{wang2020cnn} and exploiting frequency domain inconsistencies inherent to the generation process. Later work traced these cues to the upsampling operations shared across generative architectures~\citep{tan2024rethinking}. With the emergence of diffusion models~\citep{ho2020denoising, rombach2022high}, researchers developed specialized detectors targeting diffusion specific signatures. Corvi~et~al.~\citep{corvi2023detection} demonstrated that a ResNet-50 trained on LDM-generated images with careful preprocessing can generalize to unseen diffusion architectures. Rajan~et~al.~\citep{rajan2025aligned} further improved generalization by constructing aligned training pairs through VAE reconstruction, ensuring that real and synthetic images differ only in decoder artifacts. Beyond CNN-based approaches, CLIP-based methods~\citep{ojha2023towards, koutlis2024leveraging, tan2025c2p, cozzolino2024raising} leverage rich semantic representations learned from web scale data through linear probing or prompt-based adaptation. DINOv2-based approaches~\citep{guillaro2025bias, chen2025dual} similarly benefit from self-supervised pretraining, with B-Free~\citep{guillaro2025bias} proposing a bias-free training paradigm using self-conditioned reconstructions. Training-free methods such as AEROBLADE~\citep{ricker2024aeroblade} and WaRPAD~\citep{choi2025training} offer deployment flexibility by analyzing reconstruction errors or feature stability without detector-specific optimization.

Most relevant is Stay-Positive~\citep{rajan2025stay}, which constrains final layer weights to be non-negative, effectively ignoring features with negative contributions that correlate with real images. While effective at eliminating real-associated spurious features, this approach operates blindly without identifying which channels are problematic. DEAR instead leverages inpainted images to explicitly measure each channel's alignment with generated regions. Crucially, our analysis reveals that features at \emph{both} extremes of the alignment spectrum are structurally fragile. By pruning channels at both extremes, DEAR removes real-associated artifacts (e.g., compression signatures) and fake-associated artifacts (e.g., generator specific fingerprints), achieving more comprehensive robustness than single-sided constraints. See Appendix~\ref{app:ext_related} for an extended discussion.

\section{Conclusion}

We proposed DEAR to address prediction asymmetry, a phenomenon where detectors exhibit high accuracy on real images but experience significant performance drops on generated content. By utilizing inpainted images to quantify feature alignment, our framework identifies and eliminates the spurious channels driving this discrepancy. Our analysis reveals that features at both extremes of the alignment spectrum are susceptible to degradation, and our bilateral pruning strategy effectively mitigates these vulnerabilities while preserving robust forensic signals. By integrating diagnostic dissection with feature pruning and inpaint-augmented refinement of the final linear classifier, DEAR establishes a principled framework for consistent detection performance across unseen generators and post-processing scenarios.

\section*{Acknowledgements}
We thank Anirudh Sundara Rajan for kindly sharing their code and dataset, and the anonymous reviewers for their constructive feedback. This work was supported by Institute of Information \& Communication Technology Planning \& Evaluation (IITP) grant funded by the Korea government (MSIT) (No. RS-2025-02263841, Development of a Real-time Multimodal Framework for Comprehensive Deepfake Detection Incorporating Common Sense Error Analysis) and Artificial intelligence industrial convergence cluster development project funded by the Ministry of Science and ICT (MSIT, Korea) \& Gwangju Metropolitan City.

\section*{Impact Statement}

This paper advances AI-generated image detection to mitigate societal risks such as misinformation, deepfake misuse, and copyright infringement posed by increasingly powerful generative models. However, insights from this work may also inform the development of generation techniques designed to evade detection. Therefore, we emphasize the need for continuous community efforts to monitor misuse and adapt detection methods to evolving generative technologies.


\bibliography{references}

@inproceedings{bau2017network,
  title={Network dissection: Quantifying interpretability of deep visual representations},
  author={Bau, David and Zhou, Bolei and Khosla, Aditya and Oliva, Aude and Torralba, Antonio},
  booktitle={Proceedings of the IEEE conference on computer vision and pattern recognition},
  pages={6541--6549},
  year={2017}
}

@inproceedings{bau2018gan,
  title={Gan dissection: Visualizing and understanding generative adversarial networks},
  author={Bau, David and Zhu, Jun-Yan and Strobelt, Hendrik and Zhou, Bolei and Tenenbaum, Joshua B and Freeman, William T and Torralba, Antonio},
  booktitle={International Conference on Learning Representations (ICLR)},
  year={2019}
}

@inproceedings{corvi2023detection,
  title={On the detection of synthetic images generated by diffusion models},
  author={Corvi, Riccardo and Cozzolino, Davide and Zingarini, Giada and Poggi, Giovanni and Nagano, Koki and Verdoliva, Luisa},
  booktitle={ICASSP 2023-2023 IEEE International Conference on Acoustics, Speech and Signal Processing (ICASSP)},
  pages={1--5},
  year={2023},
  organization={IEEE}
}

@inproceedings{rajan2025aligned,
title={Aligned Datasets Improve Detection of Latent Diffusion-Generated Images},
author={Anirudh Sundara Rajan and Utkarsh Ojha and Jedidiah Schloesser and Yong Jae Lee},
booktitle={The Thirteenth International Conference on Learning Representations},
year={2025},
url={https://openreview.net/forum?id=doBkiqESYq}
}

@inproceedings{tan2024rethinking,
  title={Rethinking the up-sampling operations in cnn-based generative network for generalizable deepfake detection},
  author={Tan, Chuangchuang and Zhao, Yao and Wei, Shikui and Gu, Guanghua and Liu, Ping and Wei, Yunchao},
  booktitle={Proceedings of the IEEE/CVF Conference on Computer Vision and Pattern Recognition},
  pages={28130--28139},
  year={2024}
}

@inproceedings{li2025improving,
  title={Improving synthetic image detection towards generalization: An image transformation perspective},
  author={Li, Ouxiang and Cai, Jiayin and Hao, Yanbin and Jiang, Xiaolong and Hu, Yao and Feng, Fuli},
  booktitle={Proceedings of the 31st ACM SIGKDD Conference on Knowledge Discovery and Data Mining V. 1},
  pages={2405--2414},
  year={2025}
}

@inproceedings{yan2024sanity,
  title={A sanity check for ai-generated image detection},
  author={Yan, Shilin and Li, Ouxiang and Cai, Jiayin and Hao, Yanbin and Jiang, Xiaolong and Hu, Yao and Xie, Weidi},
  booktitle={International Conference on Learning Representations (ICLR)},
  year={2025}
}

@article{liang2025ferretnet,
  title={FerretNet: Efficient Synthetic Image Detection via Local Pixel Dependencies},
  author={Liang, Shuqiao and Liu, Jian and Chen, Renzhang and Guan, Quanlong},
  journal={arXiv preprint arXiv:2509.20890},
  year={2025}
}

@article{cavia2024real,
  title={Real-time deepfake detection in the real-world},
  author={Cavia, Bar and Horwitz, Eliahu and Reiss, Tal and Hoshen, Yedid},
  journal={arXiv preprint arXiv:2406.09398},
  year={2024}
}

@inproceedings{ojha2023towards,
  title={Towards universal fake image detectors that generalize across generative models},
  author={Ojha, Utkarsh and Li, Yuheng and Lee, Yong Jae},
  booktitle={Proceedings of the IEEE/CVF Conference on Computer Vision and Pattern Recognition},
  pages={24480--24489},
  year={2023}
}

@inproceedings{tan2025c2p,
  title={C2p-clip: Injecting category common prompt in clip to enhance generalization in deepfake detection},
  author={Tan, Chuangchuang and Tao, Renshuai and Liu, Huan and Gu, Guanghua and Wu, Baoyuan and Zhao, Yao and Wei, Yunchao},
  booktitle={Proceedings of the AAAI Conference on Artificial Intelligence},
  year={2025}
}

@inproceedings{cozzolino2024raising,
  title={Raising the bar of ai-generated image detection with clip},
  author={Cozzolino, Davide and Poggi, Giovanni and Corvi, Riccardo and Nie{\ss}ner, Matthias and Verdoliva, Luisa},
  booktitle={Proceedings of the IEEE/CVF Conference on Computer Vision and Pattern Recognition},
  pages={4356--4366},
  year={2024}
}

@inproceedings{koutlis2024leveraging,
  title={Leveraging representations from intermediate encoder-blocks for synthetic image detection},
  author={Koutlis, Christos and Papadopoulos, Symeon},
  booktitle={European Conference on Computer Vision},
  pages={394--411},
  year={2024},
  organization={Springer}
}

@inproceedings{yan2025effort,
  title={Effort: Efficient orthogonal modeling for generalizable ai-generated image detection},
  author={Yan, Zhiyuan and Wang, Jiangming and Wang, Zhendong and Jin, Peng and Zhang, Ke-Yue and Chen, Shen and Yao, Taiping and Ding, Shouhong and Wu, Baoyuan and Yuan, Li},
  booktitle={International Conference on Machine Learning (ICML)},
  year={2025},
}

@inproceedings{guillaro2025bias,
  title={A bias-free training paradigm for more general ai-generated image detection},
  author={Guillaro, Fabrizio and Zingarini, Giada and Usman, Ben and Sud, Avneesh and Cozzolino, Davide and Verdoliva, Luisa},
  booktitle={Proceedings of the Computer Vision and Pattern Recognition Conference},
  pages={18685--18694},
  year={2025}
}

@article{chen2025dual,
  title={Dual Data Alignment Makes AI-Generated Image Detector Easier Generalizable},
  author={Chen, Ruoxin and Xi, Junwei and Yan, Zhiyuan and Zhang, Ke-Yue and Wu, Shuang and Xie, Jingyi and Chen, Xu and Xu, Lei and Guan, Isabel and Yao, Taiping and others},
  journal={arXiv preprint arXiv:2505.14359},
  year={2025}
}

@inproceedings{chen2024drct,
  title={Drct: Diffusion reconstruction contrastive training towards universal detection of diffusion generated images},
  author={Chen, Baoying and Zeng, Jishen and Yang, Jianquan and Yang, Rui},
  booktitle={Forty-first International Conference on Machine Learning},
  year={2024}
}

@inproceedings{baraldi2024contrasting,
  title={Contrasting deepfakes diffusion via contrastive learning and global-local similarities},
  author={Baraldi, Lorenzo and Cocchi, Federico and Cornia, Marcella and Baraldi, Lorenzo and Nicolosi, Alessandro and Cucchiara, Rita},
  booktitle={European Conference on Computer Vision},
  pages={199--216},
  year={2024},
  organization={Springer}
}

@inproceedings{ricker2024aeroblade,
  title={Aeroblade: Training-free detection of latent diffusion images using autoencoder reconstruction error},
  author={Ricker, Jonas and Lukovnikov, Denis and Fischer, Asja},
  booktitle={Proceedings of the IEEE/CVF Conference on Computer Vision and Pattern Recognition},
  pages={9130--9140},
  year={2024}
}

@article{choi2025training,
  title={Training-free Detection of AI-generated images via Cropping Robustness},
  author={Choi, Sungik and Lee, Hankook and Lee, Moontae},
  journal={arXiv preprint arXiv:2511.14030},
  year={2025}
}

@article{yu2015lsun,
  title={Lsun: Construction of a large-scale image dataset using deep learning with humans in the loop},
  author={Yu, Fisher and Seff, Ari and Zhang, Yinda and Song, Shuran and Funkhouser, Thomas and Xiao, Jianxiong},
  journal={arXiv preprint arXiv:1506.03365},
  year={2015}
}

@inproceedings{lin2014microsoft,
  title={Microsoft coco: Common objects in context},
  author={Lin, Tsung-Yi and Maire, Michael and Belongie, Serge and Hays, James and Perona, Pietro and Ramanan, Deva and Doll{\'a}r, Piotr and Zitnick, C Lawrence},
  booktitle={European conference on computer vision},
  pages={740--755},
  year={2014},
  organization={Springer}
}

@inproceedings{rombach2022high,
  title={High-resolution image synthesis with latent diffusion models},
  author={Rombach, Robin and Blattmann, Andreas and Lorenz, Dominik and Esser, Patrick and Ommer, Bj{\"o}rn},
  booktitle={Proceedings of the IEEE/CVF conference on computer vision and pattern recognition},
  pages={10684--10695},
  year={2022}
}

@article{desai2021redcaps,
  title={{RedCaps}: Web-curated image-text data created by the people, for the people},
  author={Desai, Karan and Kaul, Gaurav and Aysola, Zubin and Johnson, Justin},
  journal={arXiv preprint arXiv:2111.11431},
  year={2021}
}

@misc{flux,
    author={{Black Forest Labs}},
    title={FLUX},
    year={2024},
    howpublished={\url{https://github.com/black-forest-labs/flux}},
}

@article{razzhigaev2023kandinsky,
  title={Kandinsky: an improved text-to-image synthesis with image prior and latent diffusion},
  author={Razzhigaev, Anton and Shakhmatov, Arseniy and Maltseva, Anastasia and Arkhipkin, Vladimir and Pavlov, Igor and Ryabov, Ilya and Kuts, Angelina and Panchenko, Alexander and Kuznetsov, Andrey and Dimitrov, Denis},
  journal={arXiv preprint arXiv:2310.03502},
  year={2023}
}

@article{li2024playground,
  title={Playground v2. 5: Three insights towards enhancing aesthetic quality in text-to-image generation},
  author={Li, Daiqing and Kamko, Aleks and Akhgari, Ehsan and Sabet, Ali and Xu, Linmiao and Doshi, Suhail},
  journal={arXiv preprint arXiv:2402.17245},
  year={2024}
}

@inproceedings{
chen2024pixartalpha,
title={{PixArt-$\alpha$}: Fast Training of Diffusion Transformer for Photorealistic Text-to-Image Synthesis},
author={Junsong Chen and Jincheng YU and Chongjian GE and Lewei Yao and Enze Xie and Zhongdao Wang and James Kwok and Ping Luo and Huchuan Lu and Zhenguo Li},
booktitle={ICLR},
year={2024},
}

@misc{midjourney,
    author={{Midjourney}},
    title={Midjourney},
    year={2022},
    howpublished={\url{https://www.midjourney.com/}},
}

@article{luo2023latent,
  title={Latent consistency models: Synthesizing high-resolution images with few-step inference},
  author={Luo, Simian and Tan, Yiqin and Huang, Longbo and Li, Jian and Zhao, Hang},
  journal={arXiv preprint arXiv:2310.04378},
  year={2023}
}

@inproceedings{
pernias2024wuerstchen,
title={W\"urstchen: An Efficient Architecture for Large-Scale Text-to-Image Diffusion Models},
author={Pablo Pernias and Dominic Rampas and Mats Leon Richter and Christopher Pal and Marc Aubreville},
booktitle={The Twelfth International Conference on Learning Representations},
year={2024},
url={https://openreview.net/forum?id=gU58d5QeGv}
}

@article{patil2024amused,
  title={amused: An open muse reproduction},
  author={Patil, Suraj and Berman, William and Rombach, Robin and von Platen, Patrick},
  journal={arXiv preprint arXiv:2401.01808},
  year={2024}
}

@inproceedings{ye2025loki,
  title={LOKI: A Comprehensive Synthetic Data Detection Benchmark using Large Multimodal Models},
  author={Ye, Junyan and Zhou, Baichuan and Huang, Zilong and Zhang, Junan and Bai, Tianyi and Kang, Hengrui and He, Jun and Lin, Honglin and Wang, Zihao and Wu, Tong and others},
  booktitle={International Conference on Learning Representations (ICLR)},
  year={2025}
}

@article{ho2020denoising,
  title={Denoising diffusion probabilistic models},
  author={Ho, Jonathan and Jain, Ajay and Abbeel, Pieter},
  journal={Advances in neural information processing systems},
  volume={33},
  pages={6840--6851},
  year={2020}
}

@article{lipman2022flow,
  title={Flow matching for generative modeling},
  author={Lipman, Yaron and Chen, Ricky TQ and Ben-Hamu, Heli and Nickel, Maximilian and Le, Matt},
  journal={arXiv preprint arXiv:2210.02747},
  year={2022}
}

@inproceedings{tan2024frequency,
  title={Frequency-aware deepfake detection: Improving generalizability through frequency space domain learning},
  author={Tan, Chuangchuang and Zhao, Yao and Wei, Shikui and Gu, Guanghua and Liu, Ping and Wei, Yunchao},
  booktitle={Proceedings of the AAAI Conference on Artificial Intelligence},
  year={2024}
}

@article{li2025artificial,
  title={Is Artificial Intelligence Generated Image Detection a Solved Problem?},
  author={Li, Ziqiang and Yan, Jiazhen and He, Ziwen and Zeng, Kai and Jiang, Weiwei and Xiong, Lizhi and Fu, Zhangjie},
  journal={arXiv preprint arXiv:2505.12335},
  year={2025}
}

@inproceedings{rajan2025stay,
title={Stay-Positive: A Case for Ignoring Real Image Features in Fake Image Detection},
author={Anirudh Sundara Rajan and Yong Jae Lee},
booktitle={Forty-second International Conference on Machine Learning},
year={2025},
url={https://openreview.net/forum?id=VNLmfMJi3w}
}

@inproceedings{tousi2021automatic,
  title={Automatic correction of internal units in generative neural networks},
  author={Tousi, Ali and Jeong, Haedong and Han, Jiyeon and Choi, Hwanil and Choi, Jaesik},
  booktitle={Proceedings of the IEEE/CVF Conference on Computer Vision and Pattern Recognition},
  pages={7932--7940},
  year={2021}
}

@inproceedings{wang2020cnn,
  title={CNN-generated images are surprisingly easy to spot... for now},
  author={Wang, Sheng-Yu and Wang, Oliver and Zhang, Richard and Owens, Andrew and Efros, Alexei A},
  booktitle={Proceedings of the IEEE/CVF conference on computer vision and pattern recognition},
  pages={8695--8704},
  year={2020}
}

@article{chan2001active,
  title={Active contours without edges},
  author={Chan, Tony F and Vese, Luminita A},
  journal={IEEE Transactions on image processing},
  volume={10},
  number={2},
  pages={266--277},
  year={2001},
  publisher={IEEE}
}

@article{geirhos2020shortcut,
  title={Shortcut learning in deep neural networks},
  author={Geirhos, Robert and Jacobsen, J{\"o}rn-Henrik and Michaelis, Claudio and Zemel, Richard and Brendel, Wieland and Bethge, Matthias and Wichmann, Felix A},
  journal={Nature Machine Intelligence},
  volume={2},
  number={11},
  pages={665--673},
  year={2020},
  publisher={Nature Publishing Group UK London}
}

@inproceedings{grommelt2024fake,
  title={Fake or JPEG? Revealing Common Biases in Generated Image Detection Datasets},
  author={Grommelt, Patrick and Weiss, Louis and Pfreundt, Franz-Josef and Keuper, Janis},
  booktitle={ECCV},
  year={2024}
}

@inproceedings{kashiani2025freqdebias,
title={FreqDebias: Towards Generalizable Deepfake Detection via Consistency-Driven Frequency Debiasing},
author={Kashiani, Hossein and Alipour, Niloufar and Afghah, Fatemeh},
booktitle={Proceedings of the IEEE/CVF Conference on Computer Vision and Pattern Recognition (CVPR)},
year={2025}
}

@inproceedings{ma2025specificity,
title={From Specificity to Generality: Revisiting Generalizable Artifacts in Detecting Face Deepfakes},
author={Long Ma and Zhiyuan Yan and Jin Xu and Yize Chen and Qinglang Guo and Zhen Bi and Yong Liao and Hui Lin},
booktitle={The Thirty-ninth Annual Conference on Neural Information Processing Systems},
year={2025},
url={https://openreview.net/forum?id=HCSjARBq5T}
}

@inproceedings{wang2023dire,
  title={Dire for diffusion-generated image detection},
  author={Wang, Zhendong and Bao, Jianmin and Zhou, Wengang and Wang, Weilun and Hu, Hezhen and Chen, Hong and Li, Houqiang},
  booktitle={Proceedings of the IEEE/CVF International Conference on Computer Vision},
  pages={22445--22455},
  year={2023}
}

@inproceedings{luo2024lare2,
  title={Lare\^{} 2: Latent reconstruction error based method for diffusion-generated image detection},
  author={Luo, Yunpeng and Du, Junlong and Yan, Ke and Ding, Shouhong},
  booktitle={Proceedings of the IEEE/CVF Conference on Computer Vision and Pattern Recognition},
  pages={17006--17015},
  year={2024}
}

@inproceedings{cazenavette2024fakeinversion,
  title={Fakeinversion: Learning to detect images from unseen text-to-image models by inverting stable diffusion},
  author={Cazenavette, George and Sud, Avneesh and Leung, Thomas and Usman, Ben},
  booktitle={Proceedings of the IEEE/CVF Conference on Computer Vision and Pattern Recognition},
  pages={10759--10769},
  year={2024}
}

@inproceedings{chu2025fire,
  title={Fire: Robust detection of diffusion-generated images via frequency-guided reconstruction error},
  author={Chu, Beilin and Xu, Xuan and Wang, Xin and Zhang, Yufei and You, Weike and Zhou, Linna},
  booktitle={Proceedings of the Computer Vision and Pattern Recognition Conference},
  pages={12830--12839},
  year={2025}
}

@inproceedings{liu2024fatformer,
  title={Forgery-aware adaptive transformer for generalizable synthetic image detection},
  author={Liu, Huan and Tan, Zichang and Tan, Chuangchuang and Wei, Yunchao and Wang, Jingdong and Zhao, Yao},
  booktitle={Proceedings of the IEEE/CVF Conference on Computer Vision and Pattern Recognition},
  pages={10770--10780},
  year={2024}
}

@inproceedings{wang2025sdd,
  title={Semantic Discrepancy-aware Detector for Image Forgery Identification},
  author={Wang, Ziye and Yu, Minghang and Xu, Chunyan and Cui, Zhen},
  booktitle={Proceedings of the IEEE/CVF International Conference on Computer Vision},
  pages={18388--18398},
  year={2025}
}

@inproceedings{zhou2025aigiholmes,
  title={Aigi-holmes: Towards explainable and generalizable ai-generated image detection via multimodal large language models},
  author={Zhou, Ziyin and Luo, Yunpeng and Wu, Yuanchen and Sun, Ke and Ji, Jiayi and Yan, Ke and Ding, Shouhong and Sun, Xiaoshuai and Wu, Yunsheng and Ji, Rongrong},
  booktitle={Proceedings of the IEEE/CVF International Conference on Computer Vision},
  pages={18746--18758},
  year={2025}
}

@article{zheng2024breaking,
  title={Breaking semantic artifacts for generalized ai-generated image detection},
  author={Zheng, Chende and Lin, Chenhao and Zhao, Zhengyu and Wang, Hang and Guo, Xu and Liu, Shuai and Shen, Chao},
  journal={Advances in Neural Information Processing Systems},
  volume={37},
  pages={59570--59596},
  year={2024}
}

@article{zhou2025breaking,
  title={Breaking latent prior bias in detectors for generalizable aigc image detection},
  author={Zhou, Yue and He, Xinan and Lin, KaiQing and Fan, Bing and Ding, Feng and Li, Bin},
  journal={Advances in Neural Information Processing Systems},
  volume={38},
  pages={30649--30679},
  year={2025}
}

@inproceedings{zhang2025vibnet,
  title={Towards universal ai-generated image detection by variational information bottleneck network},
  author={Zhang, Haifeng and He, Qinghui and Bi, Xiuli and Li, Weisheng and Liu, Bo and Xiao, Bin},
  booktitle={Proceedings of the Computer Vision and Pattern Recognition Conference},
  pages={23828--23837},
  year={2025}
}

@article{yuan2025mlep,
  title={MLEP: Multi-granularity Local Entropy Patterns for Generalized AI-generated Image Detection},
  author={Yuan, Lin and Li, Xiaowan and Zhang, Yan and Zhang, Jiawei and Li, Hongbo and Gao, Xinbo},
  journal={Advances in Neural Information Processing Systems},
  volume={38},
  pages={68981--69000},
  year={2025}
}

@inproceedings{fu2025pid,
  title={PiD: Generalized AI-Generated Images Detection with Pixelwise Decomposition Residuals},
  author={Fu, Xinghe and Yan, Zhiyuan and Yang, Zheng and Yao, Taiping and Zhao, Yandan and Ding, Shouhong and Li, Xi},
  booktitle={Forty-second International Conference on Machine Learning},
  year={2025}
}

@inproceedings{karageorgiou2025spai,
  title={Any-resolution ai-generated image detection by spectral learning},
  author={Karageorgiou, Dimitrios and Papadopoulos, Symeon and Kompatsiaris, Ioannis and Gavves, Efstratios},
  booktitle={Proceedings of the Computer Vision and Pattern Recognition Conference},
  pages={18706--18717},
  year={2025}
}

@inproceedings{tan2023lgrad,
  title={Learning on gradients: Generalized artifacts representation for gan-generated images detection},
  author={Tan, Chuangchuang and Zhao, Yao and Wei, Shikui and Gu, Guanghua and Wei, Yunchao},
  booktitle={Proceedings of the IEEE/CVF Conference on Computer Vision and Pattern Recognition},
  pages={12105--12114},
  year={2023}
}

@inproceedings{wang2025lota,
  title={LOTA: Bit-Planes Guided AI-Generated Image Detection},
  author={Wang, Hongsong and Cheng, Renxi and Zhang, Yang and Han, Chaolei and Gui, Jie},
  booktitle={Proceedings of the IEEE/CVF International Conference on Computer Vision},
  pages={17246--17255},
  year={2025}
}

@inproceedings{nguyen2025forensic,
  title={Forensic self-descriptions are all you need for zero-shot detection, open-set source attribution, and clustering of ai-generated images},
  author={Nguyen, Tai D and Azizpour, Aref and Stamm, Matthew C},
  booktitle={Proceedings of the Computer Vision and Pattern Recognition Conference},
  pages={3040--3050},
  year={2025}
}

@inproceedings{zhong2025beyond,
  title={Beyond Generation: A Diffusion-based Low-level Feature Extractor for Detecting AI-generated Images},
  author={Zhong, Nan and Chen, Haoyu and Xu, Yiran and Qian, Zhenxing and Zhang, Xinpeng},
  booktitle={Proceedings of the Computer Vision and Pattern Recognition Conference},
  pages={8258--8268},
  year={2025}
}

@inproceedings{chen2025forgelens,
  title={ForgeLens: Data-Efficient Forgery Focus for Generalizable Forgery Image Detection},
  author={Chen, Yingjian and Zhang, Lei and Niu, Yakun},
  booktitle={Proceedings of the IEEE/CVF International Conference on Computer Vision},
  pages={16270--16280},
  year={2025}
}

@article{zhu2023genimage,
  title={Genimage: A million-scale benchmark for detecting ai-generated image},
  author={Zhu, Mingjian and Chen, Hanting and Yan, Qiangyu and Huang, Xudong and Lin, Guanyu and Li, Wei and Tu, Zhijun and Hu, Hailin and Hu, Jie and Wang, Yunhe},
  journal={Advances in neural information processing systems},
  volume={36},
  pages={77771--77782},
  year={2023}
}

@article{lu2023seeing,
  title={Seeing is not always believing: Benchmarking human and model perception of ai-generated images},
  author={Lu, Zeyu and Huang, Di and Bai, Lei and Qu, Jingjing and Wu, Chengyue and Liu, Xihui and Ouyang, Wanli},
  journal={Advances in neural information processing systems},
  volume={36},
  pages={25435--25447},
  year={2023}
}

@article{yan2024df40,
  title={Df40: Toward next-generation deepfake detection},
  author={Yan, Zhiyuan and Yao, Taiping and Chen, Shen and Zhao, Yandan and Fu, Xinghe and Zhu, Junwei and Luo, Donghao and Wang, Chengjie and Ding, Shouhong and Wu, Yunsheng and others},
  journal={Advances in Neural Information Processing Systems},
  volume={37},
  pages={29387--29434},
  year={2024}
}

@article{pal2024semitruths,
  title={Semi-truths: A large-scale dataset of ai-augmented images for evaluating robustness of ai-generated image detectors},
  author={Pal, Anisha and Kruk, Julia and Phute, Mansi and Bhattaram, Manognya and Yang, Diyi and Chau, Duen Horng and Hoffman, Judy},
  journal={Advances in Neural Information Processing Systems},
  volume={37},
  pages={118025--118051},
  year={2024}
}

@inproceedings{park2025community,
  title={Community forensics: Using thousands of generators to train fake image detectors},
  author={Park, Jeongsoo and Owens, Andrew},
  booktitle={Proceedings of the Computer Vision and Pattern Recognition Conference},
  pages={8245--8257},
  year={2025}
}

@inproceedings{li2025rrdataset,
  title={Bridging the Gap Between Ideal and Real-world Evaluation: Benchmarking AI-Generated Image Detection in Challenging Scenarios},
  author={Li, Chunxiao and Wang, Xiaoxiao and Li, Meiling and Miao, Boming and Sun, Peng and Zhang, Yunjian and Ji, Xiangyang and Zhu, Yao},
  booktitle={Proceedings of the IEEE/CVF International Conference on Computer Vision},
  pages={20379--20389},
  year={2025}
}

@misc{fluxfill,
    author={{Black Forest Labs}},
    title={{FLUX.1} Fill},
    year={2024},
    howpublished={\url{https://huggingface.co/black-forest-labs/FLUX.1-Fill-dev}},
}

@inproceedings{podell2024sdxl,
  author    = {Podell, Dustin and English, Zion and Lacey, Kyle and Blattmann, Andreas and Dockhorn, Tim and M{\"u}ller, Jonas and Penna, Joe and Rombach, Robin},
  title     = {{SDXL}: Improving Latent Diffusion Models for High-Resolution Image Synthesis},
  booktitle = {The Twelfth International Conference on Learning Representations (ICLR)},
  year      = {2024}
}
\bibliographystyle{icml2026}

\newpage
\appendix
\onecolumn

\section{Problem Formulation and Notations}
\label{app:notation}

We address AI-generated image detection as a binary classification problem. Let $\mathcal{D} = \{(\mathbf{x}_i, y_i)\}_{i=1}^N$ denote a dataset where $\mathbf{x}_i \in \mathbb{R}^{H \times W \times C}$ represents an image and $y_i \in \{0, 1\}$ represents the label, with $y=0$ indicating real images and $y=1$ indicating AI-generated (fake) images. The objective is to train a detector that accurately estimates the probability $p(y=1|\mathbf{x})$.

We formulate the detector as a composition of two distinct modules: a deep feature backbone $f_\theta(\cdot)$ and a linear classification head $h_\phi(\cdot)$. Specifically, the backbone parameterized by $\theta$ maps an input image $\mathbf{x}$ to the feature tensor of the final convolutional layer, $\mathbf{F} = f_\theta(\mathbf{x}) \in \mathbb{R}^{D \times h \times w}$, which global average pooling reduces to a $D$-dimensional embedding $\mathbf{z} = \text{Pool}(\mathbf{F}) \in \mathbb{R}^{D}$. The classification head parameterized by $\phi = \{\mathbf{w}, b\}$ then projects this embedding to a scalar probability score via a sigmoid activation:
\begin{equation}
    p(\mathbf{x}) = h_\phi(\mathbf{z}) = \sigma(\mathbf{w}^\top \mathbf{z} + b),
\end{equation}
where $\mathbf{w} \in \mathbb{R}^{D}$ and $b \in \mathbb{R}$ denote the weight vector and bias term, respectively. This decomposition allows us to analyze and manipulate the feature extraction ($f_\theta$) and decision-making ($h_\phi$) processes independently.

\begin{table}[ht]
    \centering
    \caption{\textbf{Table of notation}}
    \label{tab:notation}
    \begin{tabular}{ll}
    \toprule 
    \textbf{Notation} & \textbf{Description} \\ 
    \midrule 
    \multicolumn{2}{l}{\textbf{Input Space \& Detector}} \\
    $\mathbf{x}$ & Input image $\mathbf{x} \in \mathbb{R}^{H \times W \times C}$. \\
    $y$ & Binary label, $y=0$ for real and $y=1$ for AI-generated (fake). \\
    $f_\theta$ & Feature backbone parameterized by $\theta$, mapping image to feature maps. \\
    $h_\phi$ & Linear classification head parameterized by $\phi = \{\mathbf{w}, b\}$. \\
    $\mathbf{w}, b$ & Weight vector and bias of the linear classifier. \\
    $\mathbf{z}$ & Feature embedding after global pooling, $\mathbf{z} = \text{Pool}(f_\theta(\mathbf{x})) \in \mathbb{R}^D$. \\
    $\mathcal{D}_{\text{train}}$ & Training dataset of real and fake image pairs. \\
    \midrule
    \multicolumn{2}{l}{\textbf{Inpainting \& Feature Maps}} \\
    $\mathbf{x}_{\text{real}}$ & Original real image used for inpainting. \\
    $\mathbf{x}_{\text{gen}}$ & AI-generated content synthesized by inpainting model. \\
    $\mathbf{x}_{\text{inpaint}}$ & Composite inpainted image with both real and generated regions. \\
    $\mathbf{M}$ & Binary inpaint mask $\mathbf{M} \in \{0, 1\}^{H \times W}$, where 1 indicates generated region. \\
    $\mathcal{D}_{\text{inpaint}}$ & Diagnostic dataset of inpainted images with ground truth masks. \\
    $\mathbf{F}$ & Feature tensor from the final conv layer, $\mathbf{F} = f_\theta(\mathbf{x}) \in \mathbb{R}^{D \times h \times w}$. \\
    $\mathbf{F}_k$ & Activation map of the $k$th channel, $\mathbf{F}_k \in \mathbb{R}^{h \times w}$. \\
    $D$ & Total number of channels in the final convolutional layer. \\
    \midrule
    \multicolumn{2}{l}{\textbf{Regional Activation Discrepancy (RAD)}} \\
    $\Omega_{\text{in}}$ & Set of spatial positions in the inpainted (generated) region. \\
    $\Omega_{\text{bg}}$ & Set of spatial positions in the background (real) region. \\
    $\mu_{\text{in}}^{(k)}$ & Mean activation of channel $k$ within the inpainted region. \\
    $\mu_{\text{bg}}^{(k)}$ & Mean activation of channel $k$ within the background region. \\
    $S_k$ & RAD value for channel $k$, defined as $S_k = \mu_{\text{in}}^{(k)} - \mu_{\text{bg}}^{(k)}$. \\
    \midrule
    \multicolumn{2}{l}{\textbf{Bilateral Pruning}} \\
    $\alpha_{\text{low}}, \alpha_{\text{high}}$ & Lower and upper percentile parameters for pruning. \\
    $\tau_{\text{low}}, \tau_{\text{high}}$ & Computed thresholds based on empirical RAD distribution. \\
    $\mathbf{m}$ & Binary pruning mask $\mathbf{m} \in \{0, 1\}^D$ indicating retained channels. \\
    $\tilde{\mathbf{F}}$ & Pruned feature map after applying mask, $\tilde{\mathbf{F}} = \mathbf{m} \odot \mathbf{F}$. \\
    \bottomrule
    \end{tabular}
\end{table}

\section{Limitations and Discussion}
\label{app:limitations}

While DEAR provides a simple yet effective approach for improving detector robustness, it exhibits certain limitations. First, our analysis focuses exclusively on the final convolutional layer before global average pooling, as it produces the feature representation used by the linear classifier. However, lower layers may capture complementary low-level artifacts such as pixel-level noise patterns or frequency domain irregularities that could further enhance detection. Extending the dissection framework to analyze and selectively prune features across multiple layers remains an important direction for future research. Second, DEAR operates as a post-hoc enhancement method that dissects and refines pretrained feature representations. An interesting alternative would be to integrate the alignment based feature selection directly into the training process, enabling the detector to learn robust features from scratch through online dissection. Exploring how to incorporate RAD based supervision during training, rather than applying it retrospectively, represents an intriguing future research direction.


\section{Empirical Evidence for Spurious Reliance}
\label{app:spurious_evidence}

We further analyze the role of spurious correlations in detector behavior through two complementary studies. The first examines the asymmetric degradation pattern under WEBP compression and shows that it matches the signature of spurious reliance on incidental curation artifacts rather than a uniform sensitivity to distribution shift. The second examines detector behavior across unseen generator families and shows that refining the classifier on the representation pruned by DEAR improves detection rather than degrading it.

\subsection{Real-Class Evidence: Asymmetric Degradation under WEBP Compression}
\label{app:webp_asymm}

Prior work~\citep{rajan2025stay, grommelt2024fake} establishes that AIGI detectors often associate incidental curation artifacts such as WEBP compression, JPEG compression, and resizing patterns with the real class. These artifacts are not inherent to the real-versus-fake distinction, so applying post-processing that introduces them to AI-generated images can cause detectors to incorrectly classify those images as real.

The degradation pattern we observe under WEBP compression is consistent with this prediction. In these experiments, WEBP compression is applied to the test images at inference time. F.Acc is averaged over the nine generators of Table~\ref{table:main_results}, and R.Acc is measured on the corresponding real set. If detector failure under post-processing reflected only limited generalization, both real accuracy and fake accuracy would be expected to decline. Instead, real accuracy remains near-perfect while fake accuracy collapses, as shown in Table~\ref{tab:webp_asymmetric}. The asymmetry is the signature of spurious reliance on compression artifacts as real-class evidence rather than uniform decline across classes.

\begin{table}[h]
\centering
\caption{\textbf{Asymmetric degradation under WEBP compression.} Corvi and Rajan maintain near-perfect real accuracy across all compression strengths while their fake accuracy collapses, a signature of spurious reliance on compression artifacts as real-class evidence.}
\label{tab:webp_asymmetric}
\renewcommand\arraystretch{1.15}
\begin{tabular}{lcccc}
\toprule
WEBP Quality & Corvi R.Acc & Corvi F.Acc & Rajan R.Acc & Rajan F.Acc \\
\midrule
0 (strongest)  & 100.00 & 0.03  & 100.00 & 0.01 \\
50             & 100.00 & 3.59  & 100.00 & 4.68 \\
90 (mildest)   & 99.60  & 33.81 & 99.83  & 37.75 \\
\bottomrule
\end{tabular}
\end{table}

When the classifier is refined on the representation pruned by DEAR, fake accuracy under WEBP compression improves substantially (Table~\ref{tab:webp_dear}). Because the base detectors already maintain near-perfect real accuracy under WEBP compression, this recovery reflects removal of spurious realness cues rather than a class-balance trade-off.

\begin{table}[h]
\centering
\caption{\textbf{DEAR mitigates WEBP-induced misclassification.} Refining the classifier on the representation pruned by DEAR improves fake accuracy by 15 to 38 percentage points relative to the base detectors across all settings.}
\label{tab:webp_dear}
\renewcommand\arraystretch{1.15}
\begin{tabular}{ccccccc}
\toprule
WEBP Quality & Corvi F.Acc & DEAR-c F.Acc & $\Delta$ & Rajan F.Acc & DEAR-r F.Acc & $\Delta$ \\
\midrule
50 & 3.59  & 23.01 & +19.42 & 4.68  & 19.67 & +14.99 \\
70 & 7.24  & 33.11 & +25.87 & 9.24  & 29.21 & +19.97 \\
90 & 33.81 & 71.48 & +37.68 & 37.75 & 66.70 & +28.95 \\
\bottomrule
\end{tabular}
\end{table}

Both the asymmetric failure pattern in Table~\ref{tab:webp_asymmetric} and the targeted recovery in Table~\ref{tab:webp_dear} support interpreting the observed failure as a consequence of detectors relying on compression artifacts as real-class evidence, rather than as a generic loss of accuracy under distribution shift.

\subsection{Fake-Class Evidence: Recovery on Unseen Generator Families}
\label{app:fake_evidence}

A complementary concern is whether the channels DEAR prunes carry genuinely spurious shortcuts rather than valid but sensitive forensic cues. We address this by examining detector behavior on unseen generator families. Prior work~\citep{yan2025effort, kashiani2025freqdebias, ma2025specificity} shows that detectors trained on a single generator family tend to learn model-specific shortcuts, including low-rank artifact subspaces, spectral biases, and generator-specific noise fingerprints, rather than universal forensic traces. A channel that captured a valid generative artifact should remain useful for detecting unseen generators sharing the underlying generative process, so its removal should hurt detection on those generators. The opposite is observed in Table~\ref{tab:unseen_gen}.

\paragraph{Test sets.}
We evaluate on three groups of generators, none of which appears in the training distribution of Corvi or Rajan. The first two are taken from the test set released by~\citet{ojha2023towards}. From it we use six non-diffusion generators, reported as an unweighted mean: the GANs BigGAN, GauGAN, ProGAN, and StarGAN, together with the two low-level vision models SAN, a second-order attention network for super-resolution, and SITD (Seeing-in-the-Dark), a low-light raw-to-RGB pipeline. We report these six jointly because none of them is diffusion-based, and we label the row accordingly rather than as \emph{GAN}, since two of the six are not adversarially trained. The second group is the deepfake subset of the same test set. The third is FLUX, the DiT-based generator already used in Table~\ref{table:main_results}.

\begin{table}[h]
\centering
\caption{\textbf{DEAR improves detection on unseen generator families.} Refining the classifier on the representation pruned by DEAR improves fake accuracy across six non-diffusion generators (BigGAN, GauGAN, ProGAN, StarGAN, SAN, SITD), a deepfake test set, and FLUX (DiT), none of which appear in the training distribution of Corvi or Rajan (both trained only on LDM-generated images). A valid forensic channel should remain useful on these unseen generators, so the recovery pattern is incompatible with the interpretation that the pruned channels carry valid cues.}
\label{tab:unseen_gen}
\renewcommand\arraystretch{1.15}
\begin{tabular}{lcccc}
\toprule
Generator Family & Corvi F.Acc & DEAR-c F.Acc & Rajan F.Acc & DEAR-r F.Acc \\
\midrule
Non-diffusion avg (6 models) & 16.83 & \textbf{64.15} (+47.32) & 7.43  & \textbf{46.35} (+38.92) \\
FLUX (DiT)         & 21.53 & \textbf{100.00} (+78.47) & 12.00 & \textbf{96.60} (+84.60) \\
Deepfake           & 84.20 & \textbf{99.50} (+15.30) & 46.93 & \textbf{97.70} (+50.77) \\
\bottomrule
\end{tabular}
\end{table}

If the removed channels encoded valid forensic information, their removal should have degraded detection on these unseen generator families. Instead, removal recovers detection across three distinct groups, including the GAN-based generators, which are architecturally distinct from the LDM training generator. We interpret a feature whose removal recovers detection across these unseen families as a spurious shortcut tied to the training generator, rather than a sensitive but valid forensic cue.

\section{Robustness of the Diagnostic Procedure to Inpainting Model Choice}
\label{app:rad_robustness}

A natural concern about DEAR's diagnostic pipeline is whether the RAD scores, and hence the pruning mask they induce, depend on the specific inpainting model used to generate the diagnostic samples. We address this concern by recomputing RAD on alternative diagnostic datasets generated by two architecturally distinct inpainting models, FLUX Fill~\citep{fluxfill} (a DiT-based inpainter) and SDXL~\citep{podell2024sdxl} (a UNet-XL inpainter), and comparing both downstream detection performance and channel-level rankings against the Stable Diffusion 1.5 diagnostic used in the main paper.

\paragraph{Detection performance across diagnostic models.}
Table~\ref{tab:rad_detection} reports the post-processed test set average AUC for DEAR-c and DEAR-r when the pruning mask is derived from each of the three diagnostic models, using the same evaluation protocol as Table~\ref{table:main_results} (bottom). All three diagnostic sources yield substantially improved detection over the base detectors. The spread across diagnostic-model choices is small (about 2.5 AUC points on DEAR-c and 1.0 AUC points on DEAR-r), and every choice clearly outperforms the corresponding baseline.

\begin{table}[h]
\centering
\caption{\textbf{Detection performance across alternative diagnostic inpainting models.} Post-processed test set average AUC for DEAR-c and DEAR-r when the pruning mask is derived from each of three diagnostic inpainting models. All three diagnostic sources yield substantial improvement over the base detectors. The spread across choices is small (about 2.5 AUC on DEAR-c and 1.0 AUC on DEAR-r).}
\label{tab:rad_detection}
\renewcommand\arraystretch{1.15}
\begin{tabular}{lcc}
\toprule
Diagnostic Model & DEAR-c AUC & DEAR-r AUC \\
\midrule
SD 1.5 & \textbf{92.25} & \textbf{97.14} \\
FLUX Fill                       & 91.19          & 96.14 \\
SDXL                            & 89.79          & 96.45 \\
\midrule
Base detector (without DEAR) & 79.23          & 92.93 \\
\bottomrule
\end{tabular}
\end{table}

\paragraph{Channel-level consistency across diagnostic models.}
To examine whether the three diagnostic models identify the same channels as carrying spurious information, we measure the agreement between RAD rankings across diagnostic models using two complementary statistics. Spearman's $\rho$ summarizes the agreement on the full channel ordering, while Jaccard similarity at the $\alpha=10\%$ pruning threshold summarizes the agreement on the actual pruning decision. Results for both Corvi and Rajan backbones are shown in Table~\ref{tab:rad_consistency}.

\begin{table}[h]
\centering
\caption{\textbf{Channel-level consistency between diagnostic inpainting models.} Spearman's $\rho$ on the full RAD ranking and Jaccard similarity on the $\alpha=10\%$ pruning decision, measured against the SD 1.5 diagnostic. Values are reported for both Corvi and Rajan backbones. Channel rankings remain well-preserved across architecturally distinct inpainting models.}
\label{tab:rad_consistency}
\renewcommand\arraystretch{1.15}
\begin{tabular}{lcc}
\toprule
Comparison & Spearman $\rho$ (Corvi / Rajan) & Jaccard $\alpha=10\%$ (Corvi / Rajan) \\
\midrule
SD 1.5 vs.\ FLUX Fill & 0.79 / 0.95 & 0.87 / 0.83 \\
SD 1.5 vs.\ SDXL      & 0.94 / 0.97 & 0.92 / 0.89 \\
\bottomrule
\end{tabular}
\end{table}

\paragraph{Interpretation.}
Spearman $\rho$ values from 0.79 to 0.97 indicate that channel rankings are well-preserved across architecturally distinct inpainting models, and Jaccard similarities at $\alpha=10\%$ from 0.83 to 0.92 indicate that most of the channels selected for pruning by one diagnostic model are also selected by the others. If RAD were biased toward features specific to SD 1.5's decoder or spectral signature, we would expect two consequences. First, channel rankings would diverge when the diagnostic inpainter is replaced by an architecturally distinct one. Second, detection performance on non-SD generators would degrade. Neither is observed. The consistent gains across the nine unseen generators (Table~\ref{table:main_results}), including FLUX (a DiT model with a 16-channel VAE), aMUSEd (a masked image model with parallel token decoding), and Midjourney (a proprietary system), provide further evidence that RAD captures generator-agnostic feature properties rather than SD 1.5-specific characteristics.

This stability is consistent with what RAD measures. RAD quantifies how a pretrained detector's internal features respond to inpainted versus surrounding regions. Because the detector's feature space is fixed by its training, channels that respond differentially to inpainted content tend to do so regardless of which inpainting model produced the inpainted region, producing similar RAD rankings. The modest variation between SD 1.5 and FLUX Fill (Jaccard from 0.83 to 0.87) reflects that different inpainting models alter slightly different image properties, while the global channel ordering remains well-preserved.

\section{Implementation Details}
\label{app:impl_detail}


\def\NoNumber#1{\STATE \textcolor{gray}{#1}}

\begin{figure}[t]
\begin{minipage}{\linewidth}
\begin{algorithm}[H]
    \caption{\textbf{DEAR: Dissect and Prune}}
    \label{alg:dear}
    \begin{algorithmic}[1]
    \STATE \textbf{Input:} Pretrained detector $(f_\theta, h_\phi)$, Diagnostic data $\mathcal{D}_{\text{inpaint}}$, Train data $\mathcal{D}_{\text{train}}$, percentiles $\alpha_{\text{low}}, \alpha_{\text{high}}$
    \STATE \textbf{Output:} Refined detector $(f_\theta, h_\phi, \mathbf{m})$

    \NoNumber{\small{\color{gray}\texttt{// Stage 1: Dissection (Compute RAD)}}}
    \STATE $S \gets \mathbf{0} \in \mathbb{R}^D$, \, $n \gets 0$
    \FOR{the first 100 batches $(\mathbf{x}, \mathbf{M})$ in $\mathcal{D}_{\text{inpaint}}$}
        \STATE Discard samples whose cropped mask is empty, skipping the batch if none remain
        \STATE $\mathbf{F} \gets f_\theta(\mathbf{x})$, \, $\widetilde{\mathbf{M}} \gets \text{Downsample}(\mathbf{M})$
        \STATE $S \gets S + \text{mean}_{\text{batch}}\big(\text{RAD}(\mathbf{F}, \widetilde{\mathbf{M}})\big)$ \, (Eqs.~\ref{eq:rad} and~\ref{eq:rad_weighted}), \, $n \gets n + 1$
    \ENDFOR
    \STATE $S \gets S / n$

    \NoNumber{\small{\color{gray}\texttt{// Stage 2: Bilateral Pruning}}}
    \STATE $\tau_{\text{low}} \gets \text{Percentile}(S, 100\,\alpha_{\text{low}})$
    \STATE $\tau_{\text{high}} \gets \text{Percentile}(S, 100\,(1 - \alpha_{\text{high}}))$
    \STATE $\mathbf{m} \gets \mathbbm{1}[\tau_{\text{low}} \leq S \leq \tau_{\text{high}}]$

    \NoNumber{\small{\color{gray}\texttt{// Stage 3: Classifier Refinement}}}
    \STATE Freeze backbone parameters $\theta$, keeping BatchNorm in eval mode
    \STATE Reinitialize classifier parameters $\phi \gets \mathbf{0}$
    \FOR{paired batches $(\mathbf{x}, y)$ in $\mathcal{D}_{\text{train}}$ and $(\mathbf{x}', y')$ in $\mathcal{D}_{\text{inpaint}}$}
        \NoNumber{\small{\color{gray}\texttt{// Inpaint label: 1 if the crop retains inpainted pixels, else 0}}}
        \NoNumber{\small{\color{gray}\texttt{// Apply pruning mask}}}
        \STATE $\tilde{\mathbf{F}} \gets \mathbf{m} \odot f_\theta(\mathbf{x})$, \, $\tilde{\mathbf{F}}' \gets \mathbf{m} \odot f_\theta(\mathbf{x}')$
        \STATE $\mathcal{L} \gets \mathcal{L}_{\text{BCE}}(h_\phi(\text{Pool}(\tilde{\mathbf{F}})), y) + \mathcal{L}_{\text{BCE}}(h_\phi(\text{Pool}(\tilde{\mathbf{F}}')), y')$
        \STATE Update $\phi$ to minimize $\mathcal{L}$
    \ENDFOR

    \STATE \textbf{return} $(f_\theta, h_\phi, \mathbf{m})$
    \end{algorithmic}
\end{algorithm}
\end{minipage}
\vspace{-0.3cm}
\end{figure}

\paragraph{Diagnostic data generation.}
We construct the diagnostic inpaint dataset using 90,000 images from the LSUN dataset. For each image, we generate a random rectangular mask with area ratio uniformly sampled from $[0.02, 0.2]$ and aspect ratio from $[0.5, 2.0]$. The mask position is randomly placed within the image boundaries. We employ the Stable Diffusion 1.5 inpainting model\footnote{\url{https://huggingface.co/stable-diffusion-v1-5/stable-diffusion-inpainting}} to synthesize content for the masked regions. We deliberately pass an \emph{empty} text prompt, so that the synthesized content is driven only by the surrounding authentic context and carries no prompt-specific semantics that a detector could latch onto.

Compositing with the hard rectangular mask would leave a step discontinuity along the mask border. Such a seam is an artifact of the compositing operation, not of the generative model, and a detector could respond to it rather than to the generative traces we intend to probe. We therefore feather the boundary: we apply a Gaussian blur with a blur factor of 33 to the mask, obtaining a soft mask $\mathbf{M}_\sigma \in [0,1]^{H \times W}$, and use $\mathbf{M}_\sigma$ both as the conditioning mask passed to the inpainting model and as the per-pixel compositing weight:
\begin{equation}
    \mathbf{x}_{\text{inpaint}} = \mathbf{M}_\sigma \odot \mathbf{x}_{\text{gen}} + (1 - \mathbf{M}_\sigma) \odot \mathbf{x}_{\text{real}}.
\end{equation}
Equation~\ref{eq:composite} in the main text states this construction with the binary mask for clarity. The hard binary mask $\mathbf{M}$ is stored separately and serves as the ground truth spatial reference for the dissection stage.

\begin{figure}[t]
    \centering
    \includegraphics[width=1.0\linewidth]{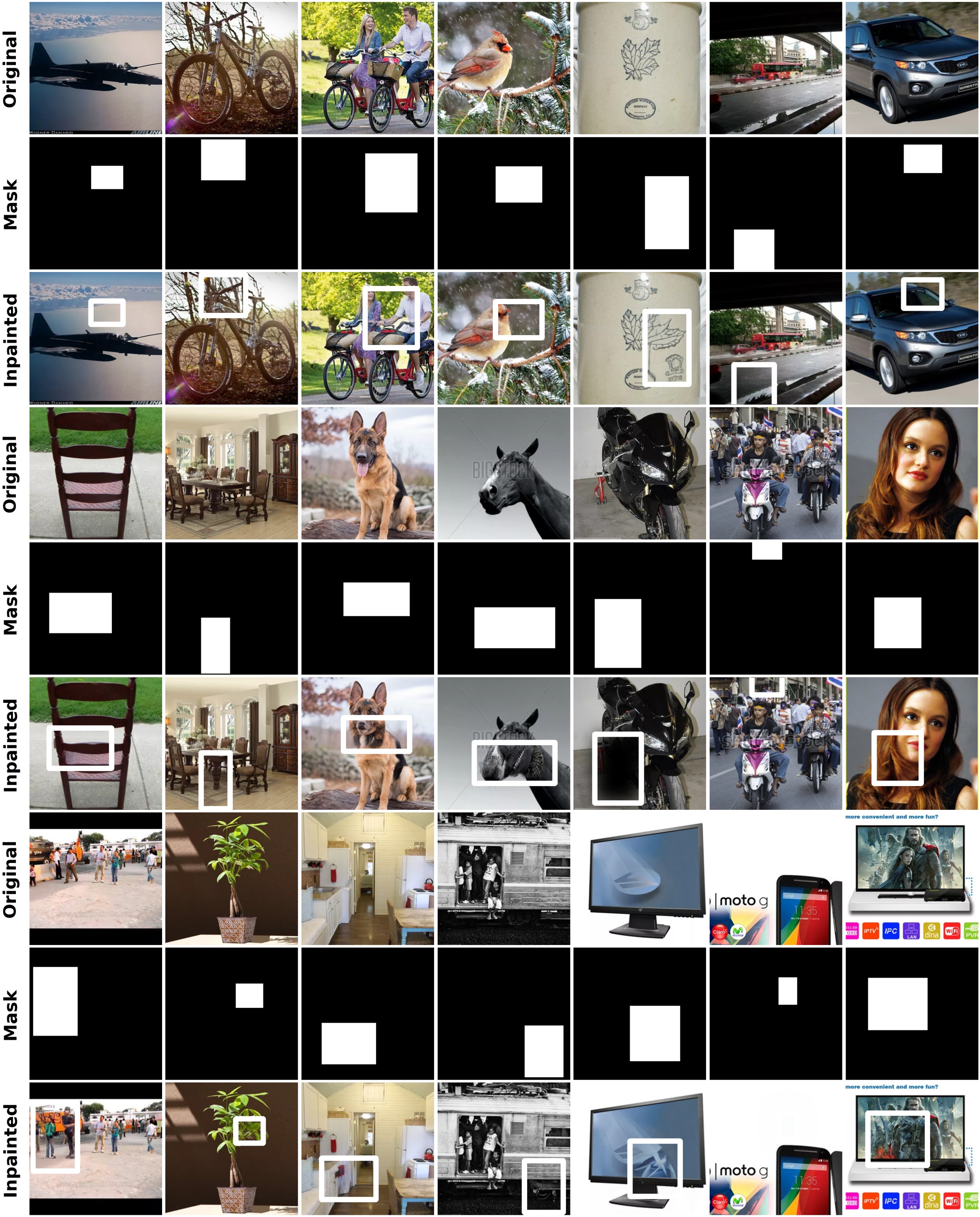}
    \caption{\textbf{Examples from the diagnostic inpaint dataset.} Each group of three rows shows: original real images (top), binary inpaint masks (middle), and resulting inpainted images with mask boundaries highlighted in white (bottom).}
    \label{fig:inpaint_samples}
\end{figure}

Inpainting is carried out at 512$\times$512 resolution: the real image and its mask are first resized to 512$\times$512, the masked region is regenerated and composited at that resolution, and the resulting composite is then resized back to the original dimensions. Figure~\ref{fig:inpaint_samples} shows representative examples from the diagnostic dataset.

\paragraph{Dissection and pruning.}
For RAD computation, we extract feature maps from the final convolutional layer before global average pooling, which contains $D=2048$ channels for the ResNet-50 backbone. Scoring uses the same input pipeline as classifier refinement: each diagnostic image passes through the training augmentations, including JPEG compression, blur, noise and color jitter, and is then randomly cropped to $96 \times 96$, with all geometric transforms applied identically to the image and its mask so that the two remain registered.

Since a $96 \times 96$ crop yields a $12 \times 12$ feature map, we downsample the binary mask onto the feature grid by bilinear interpolation, which yields a soft occupancy map $\widetilde{\mathbf{M}} \in [0,1]^{h \times w}$. The two regional means entering Eq.~\ref{eq:rad} are then computed as occupancy-weighted averages,
\begin{equation}
    \mu_{\text{in}}^{(k)} = \frac{\sum_{x} \widetilde{\mathbf{M}}(x)\,\mathbf{F}_k(x)}{\sum_{x} \widetilde{\mathbf{M}}(x)}, \quad
    \mu_{\text{bg}}^{(k)} = \frac{\sum_{x} \big(1 - \widetilde{\mathbf{M}}(x)\big)\,\mathbf{F}_k(x)}{\sum_{x} \big(1 - \widetilde{\mathbf{M}}(x)\big)},
    \label{eq:rad_weighted}
\end{equation}
where the sums run over all $hw$ spatial positions, so that positions straddling the mask boundary contribute to both terms in proportion to their occupancy.

\begin{figure}[t]
    \centering
    \includegraphics[width=1.0\linewidth]{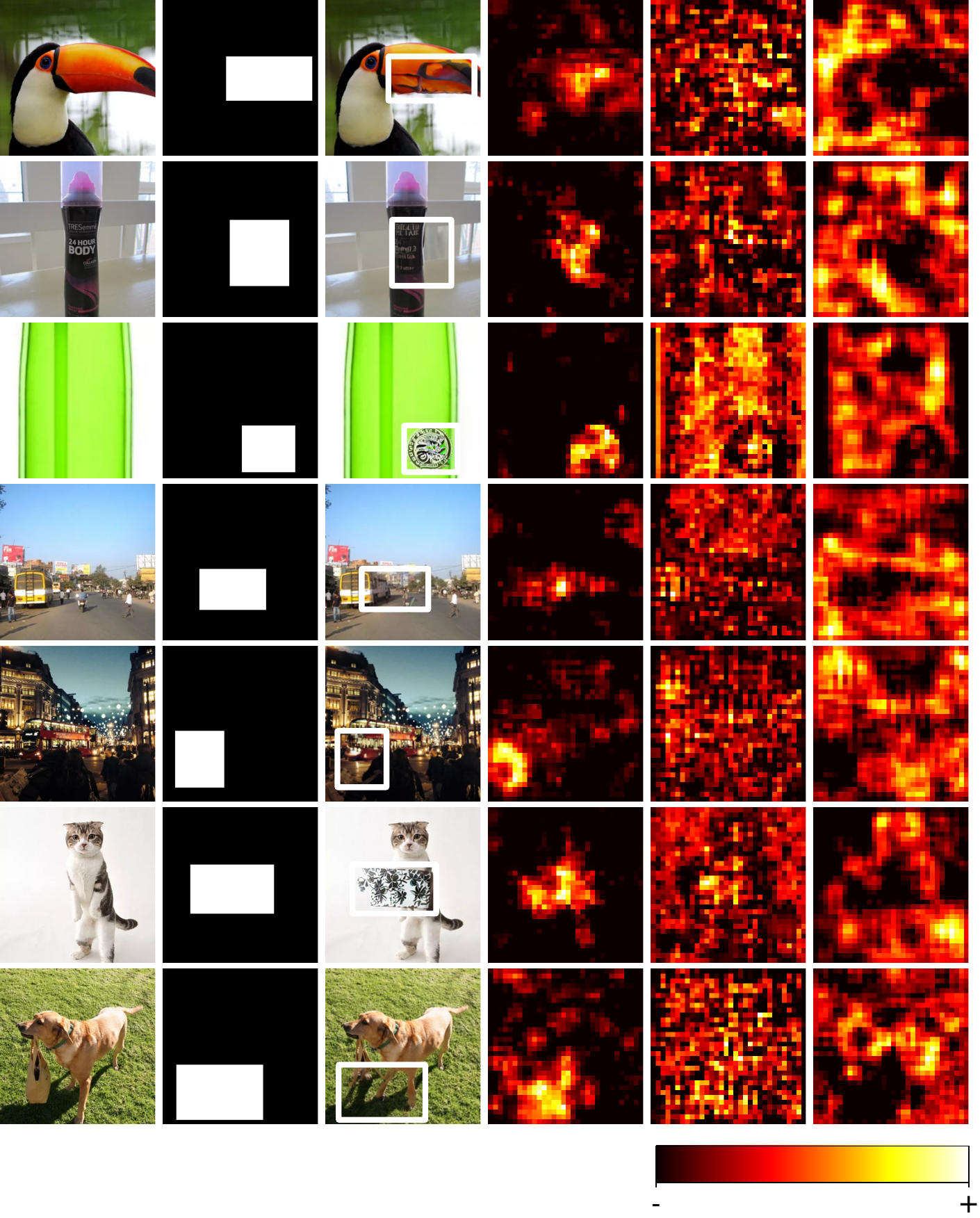}
    \caption{\textbf{Extended Feature Alignment Visualization.} Additional examples demonstrating the regional activation patterns across different images. Each row shows (from left to right): original image, inpainting mask, inpainted image, and activation maps from high, middle, and low RAD channels. Consistent with Figure~\ref{fig:feature_align}, high RAD channels selectively activate on generated regions while low RAD channels respond to real backgrounds.}
    \label{fig:feature_align_appendix}
\end{figure}

A random crop may fall entirely inside the authentic background, in which case the sample contains no inpainted region and is excluded from scoring. We accumulate per-batch means over the retained samples and average across 100 batches (approximately 6,400 loaded crops) from the diagnostic dataset. The bilateral pruning thresholds are determined by percentile parameters $\alpha_{\text{low}}$ and $\alpha_{\text{high}}$, which we tune from the set $\{0.05, 0.1, 0.2, 0.3\}$. The results reported throughout the paper use $\alpha_{\text{low}} = \alpha_{\text{high}} = 0.2$ for DEAR-c and $\alpha_{\text{low}} = \alpha_{\text{high}} = 0.3$ for DEAR-r. We apply hard gating, setting the gate value to 1 for retained channels and 0 for pruned channels. A full list of the hyperparameters is reported in Table~\ref{tab:hyperparameters}. Figure~\ref{fig:feature_align_appendix} provides additional examples of the feature alignment visualization across diverse images.

\begin{table}[htbp]
\caption{\textbf{Hyperparameters for DEAR.}}
\label{tab:hyperparameters}
\centering
\begin{tabular}{@{}llc@{}}
\toprule
\textbf{Component} & \textbf{Hyperparameter} & \textbf{Value} \\
\midrule
\multicolumn{3}{l}{\textit{Diagnostic Data Generation}} \\
& Source Images & LSUN (90,000 images) \\
& Mask Area Ratio & $[0.02, 0.2]$ \\
& Mask Aspect Ratio & $[0.5, 2.0]$ \\
& Inpainting Model & Stable Diffusion 1.5 \\
& Mask Blur Factor & 33 \\
& Processing Resolution & 512$\times$512 \\
\midrule
\multicolumn{3}{l}{\textit{Dissection (RAD Computation)}} \\
& Feature Layer & Final conv (layer4) \\
& Number of Channels ($D$) & 2048 \\
& Scoring Crop Size & 96$\times$96 \\
& Feature Map Size & 12$\times$12 \\
& Mask Downsampling & Bilinear (soft occupancy) \\
& Score Type & RAD \\
& Max Scoring Batches & 100 \\
\midrule
\multicolumn{3}{l}{\textit{Bilateral Pruning}} \\
& $\alpha_{\text{low}}, \alpha_{\text{high}}$ & Tuned from $\{0.05, 0.1, 0.2, 0.3\}$ \\
& Reported setting & 0.2 (DEAR-c) / 0.3 (DEAR-r) \\
\midrule
\multicolumn{3}{l}{\textit{Classifier Refinement}} \\
& Learning Rate & $1 \times 10^{-4}$ \\
& Optimizer & Adam ($\beta_1=0.9$) \\
& Batch Size (Main / Inpaint) & 64 / 64 \\
& Head Initialization & Zeros \\
& Objective & $\mathcal{L}_{\text{BCE}}$(main) $+$ $\mathcal{L}_{\text{BCE}}$(inpaint) \\
& Inpaint Target & 1 if crop retains mask, else 0 \\
& Early Stopping Patience & 5 epochs \\
& Early Stopping Delta & 0.001 \\
\bottomrule
\end{tabular}
\end{table}

\paragraph{Classifier refinement.}
After applying the pruning mask, we freeze all backbone parameters, keep every batch normalization layer in evaluation mode so that its running statistics are not updated, and reinitialize the final linear classifier weights and bias to zero. We train the classifier using the Adam optimizer with a learning rate of $1 \times 10^{-4}$ and $\beta_1 = 0.9$. Each optimization step draws one batch of 64 from the original training data and one batch of 64 from the diagnostic inpaint data, and minimizes the sum of the binary cross-entropy losses on the two batches. The target of a diagnostic sample is decided per crop rather than per source image: it is 1 if the crop retains any pixel of the inpainted region and 0 otherwise, so crops falling entirely inside the authentic background act as real examples. The inpaint branch therefore acts as a localized supervision signal, requiring the classifier to respond to the presence of generated pixels inside the crop rather than to image-level context. For DEAR-r, the main branch follows the aligned-pair protocol of~\citet{rajan2025aligned}: a real image and its VAE reconstruction receive identical augmentation and are concatenated into a single batch, giving an effective main batch size of 128. We apply early stopping with a patience of 5 epochs and a minimum improvement threshold of 0.001 in validation accuracy. When validation accuracy stagnates, the learning rate is reduced by a factor of 10, and training terminates when the learning rate falls below $10^{-6}$.

\paragraph{Data augmentation.}
We adopt the data augmentation pipeline established in prior work~\cite{corvi2023detection, wang2020cnn}. Augmentations are applied to the full image first, and a random 96$\times$96 crop is extracted afterwards. This crop is the input actually fed to the network. In application order, the scheme comprises random resized cropping to 256$\times$256 with aspect ratio in $[0.75, 1.33]$ ($p=0.2$) to ensure scale invariance, color jittering with brightness, contrast and saturation strength 0.4 and hue strength 0.1 ($p=0.8$), grayscale conversion ($p=0.2$), cutout of a single 48$\times$48 region ($p=0.2$), additive Gaussian noise ($p=0.2$), Gaussian blur with sigma in $[0.0, 3.0]$ ($p=0.5$), JPEG compression with quality factors sampled from $[30, 100]$ ($p=0.5$), rotation by a random multiple of $90^\circ$ ($p=1.0$), and horizontal flipping ($p=0.5$). For diagnostic inpaint samples, the geometric transforms (resized crop, rotation, flip, and the final crop) are applied identically to the image and its mask, while the photometric and degradation transforms are applied to the image only.

\paragraph{Inference.}
During inference, we do not apply any cropping or resizing to the input images. This is possible because the ResNet-50 backbone employs a spatially adaptive average pooling layer before the final classifier, allowing it to process images of arbitrary resolutions.

\paragraph{Computational resources and runtimes.}
All experiments were conducted using a single NVIDIA H200 GPU. The approximate execution times for each component are as follows:
\begin{itemize}
    \item Diagnostic inpaint data generation: 11 hours
    \item DEAR-c / DEAR-r classifier refinement: approximately 3 hours
\end{itemize}
These times are per model training instance or data generation run and may vary slightly depending on the specific environment.

\section{Experimental Setup Details}

\subsection{Details of Datasets}
\label{app:setup_data}

We describe the datasets used for training and evaluation in our experiments.

\vspace{1mm}
\noindent\textbf{Training Data.}
Following the established protocol in prior work~\citep{corvi2023detection, rajan2025aligned}, we construct our training set using real images from two widely-used sources: LSUN~\citep{yu2015lsun} and COCO~\citep{lin2014microsoft}. The LSUN dataset provides diverse indoor and outdoor scene categories, while COCO offers images with rich object annotations across various contexts. For synthetic training images, we consider two generation strategies depending on the baseline detector:
\begin{itemize}
    \item \textbf{Corvi-style training}: Fake images are generated by Latent Diffusion Models (LDM)~\citep{rombach2022high} using text prompts that correspond to the semantic content of the real images. This approach maintains content alignment through text-based conditioning.
    \item \textbf{Rajan-style training}: Fake images are produced by passing real images through the VAE encoder-decoder of LDM, creating reconstructions that preserve the exact visual content while introducing only the artifacts from the autoencoder pipeline. This alignment strategy ensures that real and synthetic pairs differ solely in decoder-induced artifacts, eliminating spurious correlations from semantic or resolution mismatches.
\end{itemize}

\vspace{1mm}
\noindent\textbf{Evaluation Data.}
For the generator-specific evaluation we adopt the test set released by~\citet{rajan2025aligned}, whose real split consists of images from the RedCaps dataset~\citep{desai2021redcaps}, which contains diverse web-crawled images with natural variations in content, resolution, and compression. The evaluation set encompasses images from multiple generative models and in-the-wild benchmarks, organized into the following categories:

\vspace{0.5mm}
\noindent\emph{Generator-specific evaluation.} We evaluate on images from nine different generative models spanning various architectures:
\begin{itemize}
    \item \textbf{Diffusion-based models}: Stable Diffusion (SD)~\citep{rombach2022high}, FLUX~\citep{flux}, Kandinsky~\citep{razzhigaev2023kandinsky}, Playground~\citep{li2024playground}, and PixArt~\citep{chen2024pixartalpha}.
    \item \textbf{Latent consistency models}: LCM~\citep{luo2023latent}, which enables fast sampling through consistency distillation.
    \item \textbf{Alternative architectures}: Wuerstchen~\citep{pernias2024wuerstchen}, which employs a three-stage cascade in which a diffusion model operates in a highly compressed latent space, and aMUSEd~\citep{patil2024amused}, a masked image model that decodes visual tokens in parallel rather than autoregressively.
    \item \textbf{Commercial systems}: Midjourney~\citep{midjourney}, a proprietary text-to-image service known for high aesthetic quality.
\end{itemize}

\vspace{0.5mm}
\noindent\emph{In-the-wild evaluation.} To assess detector performance under realistic deployment conditions, we additionally evaluate on three challenging benchmarks collected from real-world online sources:
\begin{itemize}
    \item \textbf{Chameleon}~\citep{yan2024sanity}: A curated collection of challenging AI-generated images gathered from online AI art communities such as ArtStation, Civitai, and Liblib. These images are specifically selected to be difficult for human perception and represent the current landscape of high-quality synthetic content shared publicly. We use the real and fake splits as distributed with the benchmark.
    \item \textbf{WildRF}~\citep{cavia2024real}: Images collected from social media platforms including Reddit, Facebook, and X (formerly Twitter). Fake images are retrieved using common hashtags such as \#aiart, \#aigenerated, and \#fakephoto, while real images are sourced using tags like \#photography and \#realphoto. We use both splits as released. This dataset reflects the types of images that detectors would encounter when deployed for content moderation.
    \item \textbf{LOKI}~\citep{ye2025loki}: A comprehensive synthetic data detection benchmark originally designed for evaluating large multimodal models across five modalities. We utilize its image subset, which includes diverse generation sources and specialized domains such as satellite imagery and medical images, requiring both perceptual and knowledge-based reasoning for detection. LOKI does not distribute a matched real split, so we pair it with the same RedCaps real set used for the generator-specific evaluation.
\end{itemize}

\noindent These in-the-wild benchmarks are particularly valuable because they contain images that have undergone unknown post-processing operations (compression, resizing, filtering) during upload and sharing, presenting a more realistic challenge than laboratory-controlled test sets.

\subsection{Baseline Methods}
\label{app:setup_baseline}

We compare our method against a comprehensive set of state-of-the-art AI-generated image detectors. Below we provide a brief description of each baseline, categorized by their underlying architecture and approach.

\vspace{1mm}
\noindent\textbf{CNN-based Detectors.}
\begin{itemize}
    \item \textbf{Corvi}~\citep{corvi2023detection}: A ResNet-50 based detector trained on real images from COCO and LSUN, paired with fake images generated by LDM using text prompts corresponding to the real data. The method preserves low-level forensic traces by removing the downsampling operation in the first convolutional layer and employs aggressive data augmentation to improve robustness.
    
    \item \textbf{Rajan}~\citep{rajan2025aligned}: Uses the same ResNet-50 architecture and real images as Corvi, but trains on fake images generated by VAE reconstructions of the real images rather than prompt-based generation. This alignment ensures that real and fake training samples differ only in decoder artifacts, reducing spurious correlations from content or resolution mismatches.
    
    \item \textbf{Corvi+}~\citep{rajan2025stay}: Applies the Stay-Positive algorithm to the Corvi detector. The key insight is that an image should be classified as fake only if it contains artifacts from the generative model, while the absence of such artifacts indicates a real image. The method freezes the backbone of a pre-trained Corvi detector and retrains only the last linear layer with a non-negativity constraint on weights. By clamping weights to stay positive, the detector ignores features associated with real images (which would have negative weights) and focuses exclusively on fake artifacts. This eliminates spurious correlations from post-processing artifacts (e.g., WEBP compression) that the detector may have incorrectly associated with real images.
    
    \item \textbf{Rajan+}~\citep{rajan2025stay}: Applies the same Stay-Positive algorithm to the Rajan detector. Combined with Rajan's aligned training data (VAE reconstructions), this approach achieves improved robustness to post-processing operations and better generalization to newer generators within the same family (e.g., from LDM to FLUX). The method also enables effective detection of partially inpainted images, where conventional detectors struggle due to their reliance on real image features.
    
    \item \textbf{NPR}~\citep{tan2024rethinking}: This approach feeds a ResNet-50 classifier with a residual image computed as the difference between the original image and its bilinearly interpolated version. The method exploits artifacts introduced by upsampling operations that are common across various generative architectures.
    
    \item \textbf{SAFE}~\citep{li2025improving}: A lightweight detector (1.44M parameters) that addresses training biases through three strategies: (1) replacing downsampling with crop operations to avoid artifact distortion, (2) adding ColorJitter and RandomRotation augmentation to reduce overfitting to color and semantic biases, and (3) employing patch-based random masking to enforce local awareness.
    
    \item \textbf{AIDE}~\citep{yan2024sanity}: AI-generated Image DEtector with Hybrid Features that leverages multiple experts to simultaneously extract visual artifacts and noise patterns. The method uses a ConvNeXt backbone and introduces the Chameleon dataset containing challenging AI-generated images curated from online sources.
    
    \item \textbf{FerretNet}~\citep{liang2025ferretnet}: Extracts local pixel dependency features through median filtering operations that capture neighborhood relationships based on Markov Random Field theory. The method computes residuals between the original image and its median-reconstructed version. The lightweight architecture (1.1M parameters) with depthwise separable and dilated convolutions enables efficient real-time detection.
    
    \item \textbf{LaDeDa}~\citep{cavia2024real}: Locally Aware Deepfake Detection Algorithm that operates on single 9$\times$9 image patches and outputs patch-level deepfake scores, which are then pooled to produce the final image score. The method can be distilled to Tiny-LaDeDa with only 4 convolutional layers for efficient edge deployment. 

    \item \textbf{DRCT}~\citep{chen2024drct}: Uses a ConvNeXt backbone trained with diffusion reconstruction contrastive learning. The method generates semantically aligned training pairs by reconstructing both real and fake images through DDIM inversion.
\end{itemize}

\vspace{1mm}
\noindent\textbf{CLIP-based Detectors.}
\begin{itemize}
    \item \textbf{UFD}~\citep{ojha2023towards}: Leverages the pre-trained CLIP ViT-L/14 vision encoder with a simple linear probing approach. The frozen CLIP features are shown to contain discriminative information for distinguishing real from generated images across diverse generators.
    
    \item \textbf{C2P-CLIP}~\citep{tan2025c2p}: Injects category-common prompts into the CLIP framework to enhance generalization in deepfake detection. The method learns prompt embeddings that encode real and fake concepts while keeping the vision encoder frozen.
    
    \item \textbf{RINE}~\citep{koutlis2024leveraging}: Extracts features from intermediate transformer blocks of a frozen CLIP encoder and employs a Trainable Importance Estimator module to learn adaptive weights for aggregating multi-layer representations.
    
    \item \textbf{ClipDet}~\citep{cozzolino2024raising}: Systematically explores various CLIP backbones pretrained on different datasets and proposes an ensemble approach that aggregates predictions from multiple CLIP variants to improve robustness across diverse generators.
\end{itemize}

\vspace{1mm}
\noindent\textbf{Vision Transformer-based Detectors.}
\begin{itemize}
    
    
    \item \textbf{CoDE}~\citep{baraldi2024contrasting}: Learns contrastive embeddings by training a ViT encoder to distinguish global and local image similarities. The method combines multiple classification heads including linear, SVM, and k-NN classifiers for robust detection.
\end{itemize}

\vspace{1mm}
\noindent\textbf{Training-free Detectors.}
\begin{itemize}
    \item \textbf{AEROBLADE}~\citep{ricker2024aeroblade}: Detects AI-generated images by measuring the LPIPS reconstruction error when images are passed through multiple LDM autoencoders. Generated images exhibit lower reconstruction error since they were originally produced by similar encoder-decoder pipelines.
    
    \item \textbf{WaRPAD}~\citep{choi2025training}: Measures the sensitivity of DINOv2 features to high-frequency perturbations extracted via Haar wavelet decomposition. Because DINOv2 is pretrained with RandomResizedCrop augmentation, its features are more stable under such perturbations for real images than for AI-generated ones, and the method exploits this gap.
\end{itemize}

For fair comparison, we follow the official implementations and pretrained checkpoints released by the authors. 


\section{Extended Related Work}
\label{app:ext_related}

We discuss recent work on AI-generated image detection across six themes.

\paragraph{Reconstruction-Error Detectors.}
One group of detectors exploits a generative model's own reconstruction error as a forensic signal. DIRE~\citep{wang2023dire} introduces this formulation by passing test images through a pretrained DDIM (encode then decode) and measuring the per-pixel discrepancy, exploiting the observation that AI-generated images, which lie on the diffusion model's image distribution, reconstruct more accurately than authentic photographs. AEROBLADE~\citep{ricker2024aeroblade} extends the idea to a training-free setting by computing LPIPS-based reconstruction error across multiple LDM autoencoders. LaRE$^{2}$~\citep{luo2024lare2} refines DIRE by operating on the \emph{latent} reconstruction error and using it to guide CLIP-feature refinement, while FakeInversion~\citep{cazenavette2024fakeinversion} replaces the reconstruction step with a Stable Diffusion inversion and learns features that transfer to unseen text-to-image models. FIRE~\citep{chu2025fire} sharpens the reconstruction signal by decomposing it into frequency bands and weighting the bands that survive post-processing.

\paragraph{CLIP / VLM / MLLM Detectors.}
Another group of detectors leverages frozen vision--language backbones, exploiting their broad semantic priors. UFD~\citep{ojha2023towards}, ClipDet~\citep{cozzolino2024raising}, C2P-CLIP~\citep{tan2025c2p}, and RINE~\citep{koutlis2024leveraging} apply linear probing or prompt-style adaptation on CLIP features. FatFormer~\citep{liu2024fatformer} inserts forgery-aware adapters that fuse image- and frequency-domain forgery traces with a language-guided alignment branch, and ForgeLens~\citep{chen2025forgelens} similarly attaches a lightweight forgery-focus module to CLIP-ViT that is trained to suppress forgery-irrelevant CLIP features. SDD~\citep{wang2025sdd} introduces a token sampling stage that emphasizes regions where CLIP semantics and forgery cues diverge. AIGI-Holmes~\citep{zhou2025aigiholmes} moves further toward multimodal large language models, combining instruction tuning and direct preference optimization to produce an MLLM-based detector with natural-language explanations.

\paragraph{Bias and Shortcut Removal.}
Several recent works address spurious correlations in AIGI detection through different mechanisms. Stay-Positive~\citep{rajan2025stay} retrains the last linear layer of a pretrained detector under a non-negativity constraint, effectively suppressing real-aligned features while leaving fake-aligned features untouched. Breaking Semantic Artifacts~\citep{zheng2024breaking} attacks the same problem from the data side, applying patch shuffling during training to discourage the detector from relying on object-level cues. Breaking Latent Prior Bias~\citep{zhou2025breaking} identifies a different shortcut, the latent-noise prior of diffusion generators, and mitigates it via on-manifold adversarial training, releasing a GenImage++ test set that re-exposes the underlying generative artifacts. VIB-Net~\citep{zhang2025vibnet} compresses the CLIP feature representation under a variational information-bottleneck objective, aiming to retain only the dimensions that carry forgery cues. MLEP~\citep{yuan2025mlep} replaces the input with multi-granularity local entropy patterns to suppress content bias, and PiD~\citep{fu2025pid} decomposes the input image into semantic and non-semantic residual components and trains the detector on the residual alone. DEAR shares the diagnosis with these works but addresses it through direct channel-level pruning, using inpaint masks to identify channels at both extremes of the alignment spectrum.

\paragraph{Spectral and Low-Level Artifact Representations.}
A complementary group of detectors relies not on the network's learned features but on transformed inputs that expose generative artifacts directly. In the spectral direction, SPAI~\citep{karageorgiou2025spai} learns a self-supervised masked spectral model on real images and detects AI-generated content via a spectral reconstruction similarity score, handling arbitrary input resolutions through spectral context attention. The closely related frequency-masking approach of~\citet{tan2024frequency} pursues the same idea. Along a different low-level axis, LGrad~\citep{tan2023lgrad} replaces the input with image gradients computed through a frozen CNN, learning detectors on these gradient maps rather than raw pixels, and NPR~\citep{tan2024rethinking} extends the same intuition by feeding bilinear-residual representations as inputs. LOTA~\citep{wang2025lota} decomposes inputs into bit-plane noise channels and selects gradient-rich patches for classification.

\paragraph{Natural-Image-Distribution Detectors.}
A distinct paradigm trains the detector exclusively on real images and treats AI-generated content as out-of-distribution relative to the learned real-image distribution. Forensic Self-Descriptions~\citep{nguyen2025forensic} learns a bank of predictive filters whose residuals form a per-image forensic ``description'', and AI-generated images are flagged in a zero-shot manner via their deviation from the distribution of these descriptions. Beyond Generation~\citep{zhong2025beyond} repurposes a diffusion model as a denoiser inside a self-supervised pretext and models the resulting low-level feature distribution of real photographs, against which test images are compared. SPAI's spectral reconstruction similarity score~\citep{karageorgiou2025spai} also falls in this paradigm, although its defining signal is spectral.

\paragraph{Benchmarks.}
Several recent benchmarks evaluate AIGI detectors at scale and under realistic conditions. GenImage~\citep{zhu2023genimage} provides a million-scale real--fake pair benchmark spanning eight generators and has been widely adopted for cross-generator evaluation. Sentry-Image~/~Fake2M~\citep{lu2023seeing} pairs detector evaluation with a human-perception study, exposing the gap between machine and human reliability. DF40~\citep{yan2024df40} curates forty deepfake techniques, primarily face-focused, for next-generation detection evaluation. Semi-Truths~\citep{pal2024semitruths} and AIGIBench~\citep{li2025artificial} explicitly target detector robustness, with Semi-Truths releasing 1.3M AI-augmented images under controlled perturbations and AIGIBench benchmarking eleven detectors across twenty-three fake subsets and real-world social-media samples. Community Forensics~\citep{park2025community} pushes generator diversity to thousands, supplying 2.7M images from over four thousand generators. RRDataset~\citep{li2025rrdataset} measures real-world robustness through transmission and re-digitization pipelines, evaluating seventeen detectors and ten VLMs against real-world post-upload distortions. Our wild benchmarks (Chameleon~\citep{yan2024sanity}, WildRF~\citep{cavia2024real}, LOKI~\citep{ye2025loki}) fall into this category.

\section{Additional Results}
\label{app:add_results}

We provide extended ablation studies comparing DEAR variants against baseline methods and their extensions. In addition to the baselines described in Appendix~\ref{app:setup_baseline}, we introduce two additional variants to investigate whether simply augmenting training data with inpainted images can achieve similar benefits to DEAR:

\begin{itemize}
    \item \textbf{Corvi-inpaint}: Trains a ResNet-50 detector from scratch using the standard Corvi training protocol, but augments the training set with our diagnostic inpainted images as additional fake samples.
    \item \textbf{Rajan-inpaint}: Similarly trains from scratch with the Rajan protocol, augmented with inpainted images.
\end{itemize}

\noindent These variants test whether the performance gains of DEAR stem merely from exposure to inpainted data during training, or whether the dissection and pruning mechanism provides orthogonal benefits.

\begin{figure}[ht]
    \centering
    \includegraphics[width=1.0\linewidth]{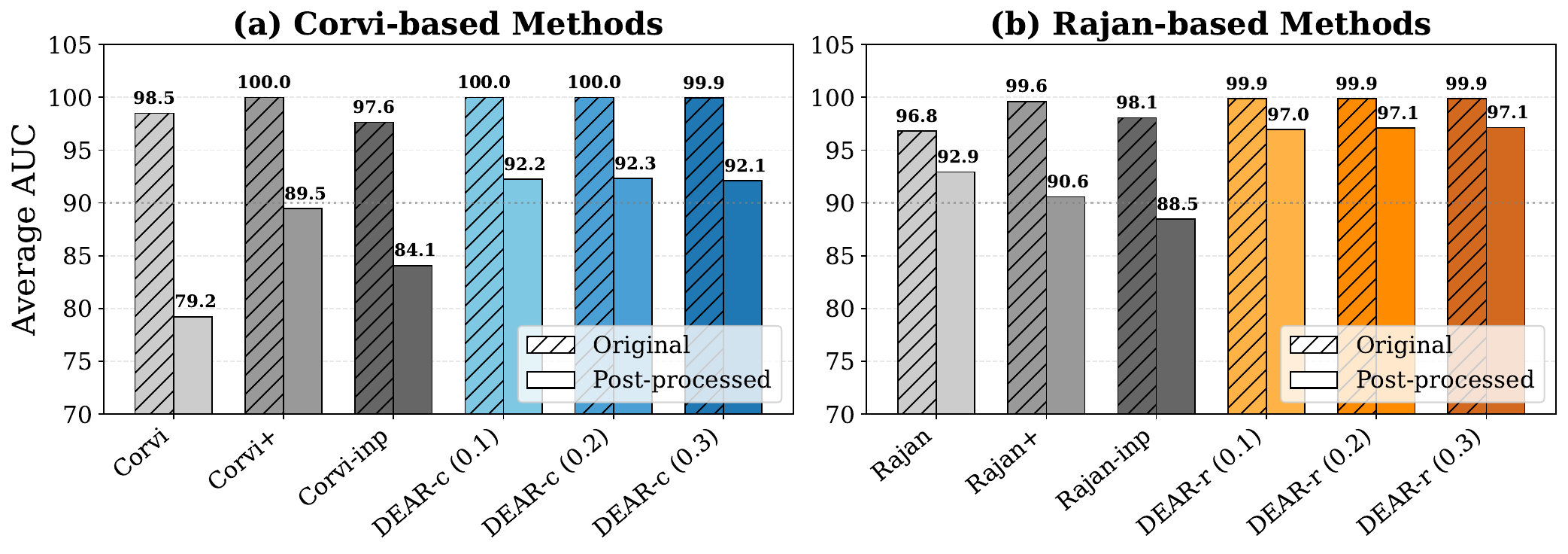}
    \caption{\textbf{Average AUC comparison across generators.} We compare baseline detectors (Corvi, Rajan), their Stay-Positive variants (Corvi+, Rajan+), inpaint-trained variants (Corvi-inpaint, Rajan-inpaint), and DEAR with varying pruning ratios ($\alpha \in \{0.1, 0.2, 0.3\}$, three of the four values in the tuning range $\{0.05, 0.1, 0.2, 0.3\}$). Hatched bars indicate performance on original images. Solid bars indicate performance on post-processed images. DEAR consistently achieves the highest average AUC, particularly under post-processing where the performance gap is most pronounced.}
    \label{fig:appendix_comparison_avg}
\end{figure}

\paragraph{Analysis of results.}
Figure~\ref{fig:appendix_comparison_avg} summarizes the average AUC across all nine generators. Several observations emerge from this comparison. First, DEAR variants consistently match or outperform all baselines on both original and post-processed images, with the performance gap being most pronounced under post-processing. Second, and perhaps more interestingly, Corvi-inpaint and Rajan-inpaint show limited improvement over their base detectors despite being trained with inpainted data. In some cases, such as Rajan-inpaint on post-processed images, performance actually degrades compared to the original Rajan detector. This suggests that naively augmenting training data with inpainted images does not effectively teach the detector to focus on robust features. In contrast, DEAR's explicit dissection and pruning mechanism successfully identifies and removes the problematic channels, achieving substantial gains without retraining the backbone from scratch.

\begin{figure}[ht]
    \centering
    \includegraphics[width=1.0\linewidth]{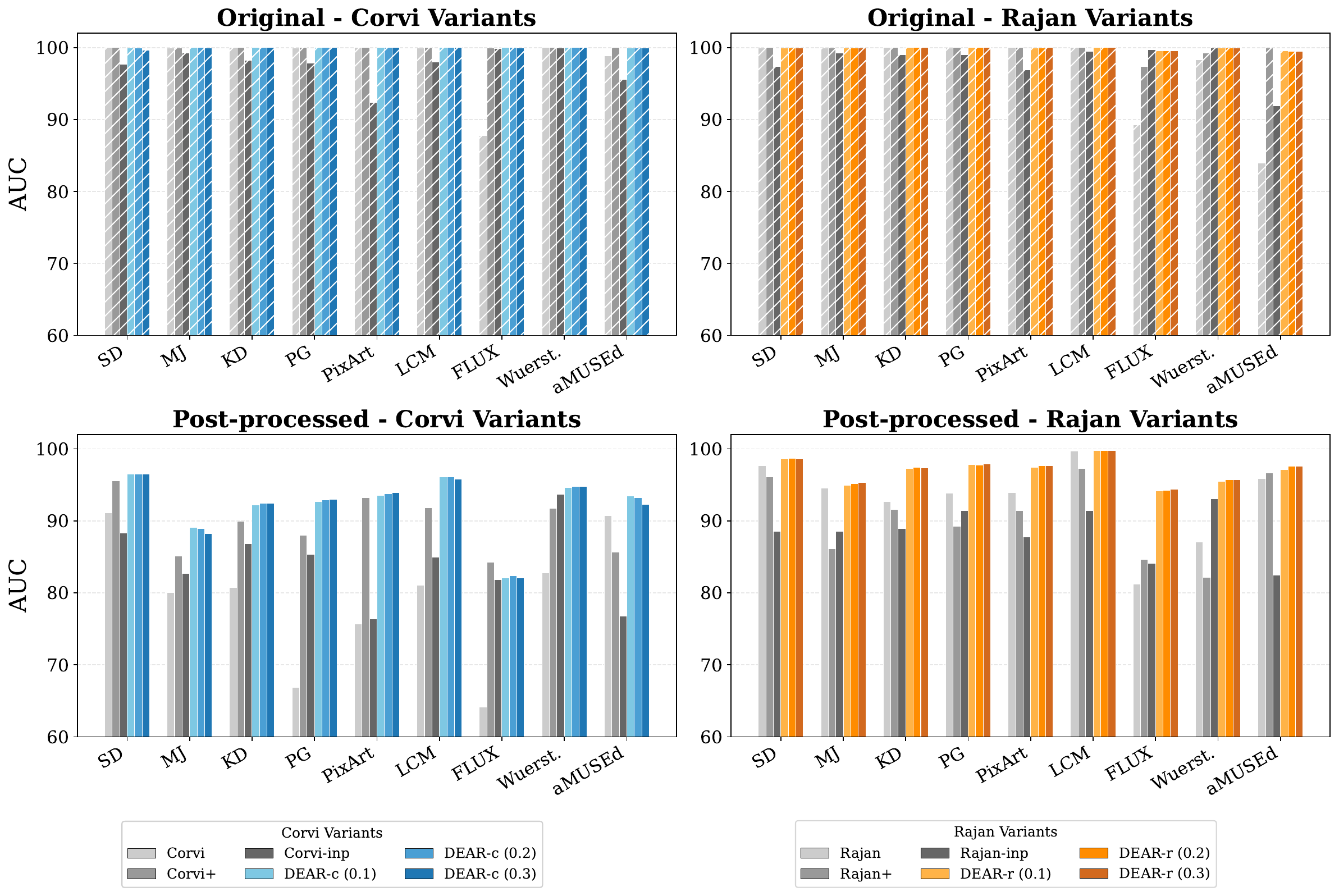}
    \caption{\textbf{Per-generator AUC comparison.} Detailed breakdown of detection performance across nine generators (SD, MJ, KD, PG, PixArt, LCM, FLUX, Wuerstchen, aMUSEd). Top row shows results on original images. Bottom row shows results on post-processed images. DEAR variants (blue for Corvi-based, orange for Rajan-based) demonstrate superior robustness.}
    \label{fig:appendix_comparison}
\end{figure}

Figure~\ref{fig:appendix_comparison} provides a per-generator breakdown of these results. The detailed view reveals that DEAR's improvements are consistent across generators. We use symmetric pruning with $\alpha_{\text{low}} = \alpha_{\text{high}} = \alpha$ throughout our experiments, and tune $\alpha$ over $\{0.05, 0.1, 0.2, 0.3\}$. The two figures plot the three largest of these values. The stability across pruning ratios further demonstrates the robustness of our approach to hyperparameter choices.

\end{document}